\pdfoutput=1
\documentclass{article}
\usepackage[T1]{fontenc}

\AtBeginDocument{%
  }

\usepackage{hyperref}
\usepackage[numbers]{natbib}
\usepackage{amsmath, amsfonts}
\usepackage{color}
\usepackage{graphicx}
\usepackage{subfigure}
\usepackage{titlesec}
\usepackage{indentfirst}
\usepackage{colortbl}
\usepackage{xcolor}
\usepackage{booktabs}
\usepackage{cleveref}
\usepackage{algorithm}
\usepackage{algpseudocode}
\usepackage{textcomp}
\usepackage{booktabs}
\usepackage{caption}
\usepackage{tikz}
\usetikzlibrary{shapes}
\usepackage{authblk}
\usepackage{units}
\usepackage{amsmath} %
\usepackage{relsize} %

\newcommand\blfootnote[1]{%
  \begingroup
  \renewcommand\thefootnote{}\footnote{#1}%
  \addtocounter{footnote}{-1}%
  \endgroup
}

\begin{document}

\title{ModeConv: A Novel Convolution for Distinguishing Anomalous and Normal Structural Behavior}

\author{Melanie Schaller\thanks{\{schaller,schloer,hotho\}@informatik.uni-wuerzburg.de}}%
\author{Daniel Schlör}%
\author{Andreas Hotho}%

\affil{Center for Artificial Intelligence and Data Science (CAIDAS),\newline Chair for Data Science (Computer Science X),\newline University of Würzburg}

\date{}

\maketitle
\blfootnote{The authors acknowledge the funding support from the German Ministry for Digital and Transport (BMDV).}
\begin{abstract}
External influences such as traffic and environmental factors induce vibrations in structures, leading to material degradation over time. These vibrations result in cracks due to the material's lack of plasticity compromising structural integrity. Detecting such damage requires the installation of vibration sensors to capture the internal dynamics. However, distinguishing relevant eigenmodes from external noise necessitates the use of Deep Learning models. The detection of changes in eigenmodes can be used to anticipate these shifts in material properties and to discern between normal and anomalous structural behavior. Eigenmodes, representing characteristic vibration patterns, provide insights into structural dynamics and deviations from expected states. Thus, we propose ModeConv to automatically capture and analyze changes in eigenmodes, facilitating effective anomaly detection in structures and material properties. In the conducted experiments, ModeConv demonstrates computational efficiency improvements, resulting in reduced runtime for model calculations. The novel ModeConv neural network layer is tailored for temporal graph neural networks, in which every node represents one sensor. ModeConv employs a singular value decomposition based convolutional filter design for complex numbers and leverages modal transformation in lieu of Fourier or Laplace transformations in spectral graph convolutions. We include a mathematical complexity analysis illustrating the runtime reduction. 
\end{abstract}

\section{Introduction}
The condition of materials or structures, such as bridges or buildings, may deteriorate over time due to various factors. To prevent failures and ensure safety, continuous monitoring of their condition is necessary. External forces, as well as environmental effects, can excite structures and materials, causing them to vibrate. These vibrations, combined with factors like deadweight \cite{bagavathiappan2013infrared, abu2002predictive, mehta2001building}, can lead to cracks and damages, ultimately compromising structural integrity.

Traditional methods for condition monitoring, such as visual inspection~\cite{graybeal2002visual, agdas2016comparison, davies2012handbook, caetano2011vision}, have limitations in detecting internal damage. To overcome these limitations, the installation of vibration sensors, such as strain gauges or accelerometers~\cite{choi2008structural, yang2015condition, dos2014overview}, allows to capture the vibration behavior of the structure or material. By analyzing the recorded sensor data, it is possible to gain insights into the global vibration behavior and detect changes in the material's condition.

However, the sensor recordings under real-world conditions contain frequencies of external forced vibrations, which can overlap with disturbance variables~\cite{fahy2007sound} like environmental conditions, ambient vibrations~\cite{ivanovic2000ambient}, or external mechanical forces and the natural frequencies of the material. These Natural frequencies are inherent frequencies at which a structure naturally vibrates~\cite{peeters2001stochastic, bachmann1995vibration, beards1996structural} and depend on the structural properties, such as mass, stiffness, and geometry~\cite{schommer2017damage, gavin2014structural}. These overlapping frequencies make it difficult to accurately identify~\cite{vulli2009time, jin2016damage} and extract the material's characteristic frequencies and modes from the sensor data.

\begin{figure}[htbp]%
\centering
\includegraphics[width=1\textwidth]{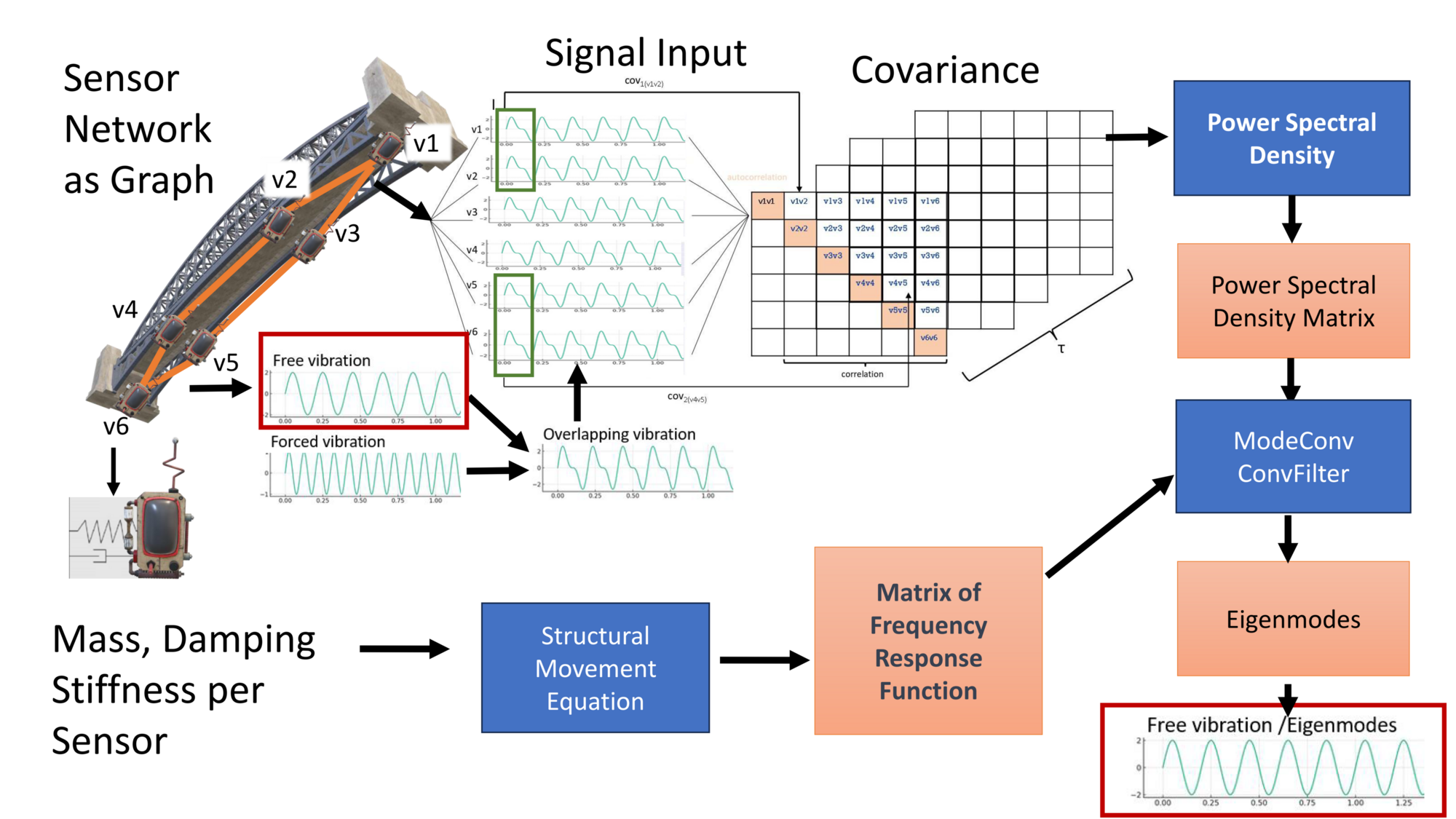}
\caption{Visualization of ModeConv Concept: In the upper part starting from the left to the right the installed Sensor Network is showcased as a graph with sensors as nodes and the connections between them as edges; all nodes get the overlapping signal of free vibrations and external vibrations as input signal (showcased with two sinus curves in different frequencies here); these mixed signals are used as input for the covariance matrix; the covariance of signals is further used to calculate the Power Spectral Density (PSD) and the PSD matrix. In the lower part of the image starting from the left to the right again, the sensornodes are represented as mass points with a certain stiffness and damping, these parameters are used as input for the equation of motion in structural dynamics to calculate the matrix of frequency response functions; Then these two matrices are used together with the PSD matrix as input for the ModeConv Convolutional filter for complex numbers with the imaginary part as first dimension and the real part as second dimension. This filter is designed to automatically learn the Eigenmodes, which are the inherent free vibrations and filters out the external effects, that are not relevant to detect anomalies in the the structural behaviour.}\label{Layer}
\end{figure}

To address the issue of overlapping frequencies and effectively capture the distinctive frequencies and modes of materials, we utilize a deep learning model with ModeConv layers.  These layers can be coupled with any sort of Graph Neural Network (GNN), in which every sensor is represented as node and the connections between them as edges, allowing us to effectively capture and separate the relevant information from the sensor recordings (see Fig. \ref{Layer}).

The ModeConv approach enables us to leverage the changes in eigenmodes for the specific task of anomaly detection\cite{hawkins1980anomaly, BlazquezGarcia2021, cateni2008outlier, xi2008outlier, papadimitriou2003cross}. In this paper, an anomaly refers to an unusual or atypical behavior or condition observed in the data collected from sensors deployed in a structure. Anomalies are deviations from the expected or normal patterns in the sensor readings. The goal of anomaly detection in the monitoring of structures is to identify these deviations, which may indicate potential structural issues, damage, or abnormal conditions within the monitored system. Anomalies in sensor data could be indicative of structural faults, damage, environmental changes, or other factors that require attention and further investigation to ensure the integrity and safety of the monitored structure.

The objectives of this paper are twofold. Firstly, we aim to develop a method to automatically detect changes in the eigenmodes of structures, allowing for early detection of damages and distinguishing between normal and anomalous structural behaviour. Secondly, we seek to reduce the computational cost of analyzing sensor data without compromising accuracy, paving the way for real-time monitoring applications in the future.

ModeConv, the novel convolutional layer proposed in this work, consists of three major building blocks: 
(1) The signal block models the covariance ~\cite{stein2005space} of the sensor data, capturing the similarity between different measurements as well as initializing and updating the edge weights ~\cite{Gori2005ANM,Yu_2018,wu2020comprehensive} of the graph. This block furthermore enables the analysis of the structural behavior by examining the correlation patterns. (2) The Partial Differential Equation (PDE) block covers the physical dynamics of the structure based on the equation of motion ~\cite{patil2000decoupled, siedziako2017enhanced, buchholdt1997structural}. It accurately represents the behavior of the structure by considering factors such as mass and stiffness ~\cite{caetano2011vision, schommer2017damage, doetsch2013handbuch}. By incorporating this physical knowledge, we can gain a deeper understanding of the structural response. (3) The convolutional block combines the information from the signal block and the PDE block to complete the modal transformation. This process involves extracting the material-typical frequencies and the corresponding natural modes. By aggregating the data through modal transformation, we obtain a comprehensive representation of the global vibration behavior of the structure. Building upon the concept of mode decomposition, we propose a novel graph convolution operation with a complex convolutional filter. To achieve this, we combine mode decomposition, learnable weights, and real-imaginary factorization. In signal processing, signals are often represented as complex numbers ~\cite{tatzko2019use, michel2006artificial, gavili2017shift}, where the real part corresponds to the signal's amplitude, and the imaginary part represents the phase shift ~\cite{servin2009general}. This representation allows for a concise description of signals. By decomposing the signal data into real and imaginary components, these can be passed as separate channels into the two-dimensional convolutional filter. By employing ModeConv, these specialized filters can be learned to capture important patterns or features in both parts.

Moreover, ModeConv achieves a reduction in computational cost by efficiently processing only the relevant information. In contrast to previous approaches, that rely on the computation of the symmetric normalized Laplace matrix with Chebyshev polynomials \cite{defferrard2017convolutional}, with a time complexity depending on the filter size $K$, the Singular Value Decomposition (SVD) \cite{stewart1993early, henry19928, wall2003singular} in ModeConv reduces the matrix multiplication to the number of extracted eigenmodes. This results in significant time savings, especially when large values of $K$ are required. 

Furthermore, ModeConv demonstrates its capability to handle unbalanced data in the two unbalanced, large-scale datasets utilized in this paper. This further enhances its practical applicability and robustness in real-world condition monitoring of structures.

Recent studies also suggest, that further improvements are necessary to process larger datasets ~\cite{https://doi.org/10.48550/arxiv.1902.07153} and to find suitable representations for real-world graphs ~\cite{grady2010discrete, ray2013graph, krim2015geometric, cvetkovic1980spectra}. In order to tackle these challenges, we employ large-scale, real-world sensornet data and focus on addressing the computational costs. 

In summary, the novel ModeConv graph convolutional layer brings forth valuable contributions to the field of structural health monitoring. It captures changes in eigenmodes, enhances accuracy even for imbalanced distributions, and reduces computational costs by processing only the essential structural information. These advancements pave the way for more effective and efficient structural health monitoring systems, ultimately improving the safety and reliability of structures. 

\section{Related Work}
The related work is delineated into three principal subsections. The initial division pertains to the model categorization of ModeConv within the framework of geometric deep learning taxonomy. Subsequently, the second subsection elucidates the benchmarking models employed for comparative analysis in the experimental section, while the third subsection delves into the pertinent literature concerning the foundational physical principles underlying the ModeConv approach.

\subsection{Model categorisation within GNN taxonomy}
Graph Convolutional Networks (GCNs) have gained significant attention in the field of graph representation learning \cite{zhang2019graph, wu2019simplifying, chen2020simple, gao2018large, manessi2020dynamic, zhang2018graph, pei2020geom} and have been widely applied to various domains. Traditional GCNs ~\cite{https://doi.org/10.48550/arxiv.1609.02907}, perform convolutions directly on the graph structure using neighborhood aggregation, but are computational complex. To tackle the computational complexity issue, several approaches have been proposed. 

One line of research focuses on approximating graph convolutions using low-rank approximations (e.g. ~\cite{defferrard2017convolutional}). Spectral-based architectures are built on multiple hidden layers, each one performing spectral graph convolutions defined from a graph signal processing point of view. It provides a notion of frequency and the graph Fourier transform, allowing for filtering in the spectral domain. As a consequence, a graph convolutional layer can be written as a sum of filtered signals followed by an activation function. Each filter is defined in the spectral domain by making use of the eigenvalue decomposition of a graph Laplacian. 

Following the proposed taxonomy of Wu et al. ~\cite{wu2020comprehensive} to select representative benchmarking strategies, the second line of convolutions can be defined as the line of spatio-temporal graphs, that share the same idea of information propagation with recurrent Graph Neural Networks. The third group consists of hybrid models, that use both.

Our approach differs from previous methods by incorporating a novel complex convolutional filter design based on the Singular Value Decomposition (SVD). Speaking by the taxonomy of Wu et al. ~\cite{wu2020comprehensive}, we thus formulate a novel group of GNNs, namely the modal graph convolutional networks, that neither represent spectral graph convolutions, nor spatio-temporal graph networks or their coupled versions. 

\subsection{Benchmarking models}
Spatial dependencies, also known as inter-variable relationships, significantly influence a model's forecasting capability \cite{Jin_2023}. When dealing with time series data accompanied by graph structures that illustrate the intensity of connections between time series, existing papers commonly utilize (1) spectral Graph Neural Networks (GNNs), (2) spatial GNNs, or (3) a coupled approach incorporating both, to capture and model these spatial dependencies. These methodologies are rooted in the principles of graph signal processing. For our study, we select benchmark models based on these grouping. Including their popularity and usage within the research community, we choose at least one model out of these groups, to compare it to the performance of the ModeConv model. Besides these three groups, we also decided to use the proposed GraphCON wrapper \cite{rusch2022graphcoupled} as fourth benchmarking section, because it is based on the idea of using nodes as oscillators, which comes closest to the ModeConv idea to model the nodes as positions in a bending wave.

Within the first group of spectral Graph Neural Networks, early papers predominantly employed ChebConv to approximate graph convolution using Chebyshev polynomials. This approach was utilized to model inter-variable dependence. As ChebConv \cite{https://doi.org/10.48550/arxiv.1606.09375} is still a widely-used model for spectral graph convolution, it is often recommended as a reference method for efficient graph convolution and has demonstrated success in numerous applications. StemGNN \cite{https://doi.org/10.48550/arxiv.2103.07719} further introduces spectral-temporal graph neural networks that extract intricate time series patterns by utilizing ChebConv and frequency-domain convolution. Specifically, it captures inter-series correlations and temporal dependencies in the spectral domain using Graph Fourier Transform (GFT) for inter-series correlations and Discrete Fourier Transform (DFT) for temporal dependencies in an end-to-end framework. Although there have been challenges in reproducing the results of StemGNN due to differences in the official implementation, it is still considered a relevant model for spatio-temporal graph networks and thus mentioned here, but the results are excluded from the benchmarking experiments.

Within the second group of spatial GNNs \cite{9046288}, one line of research has been modeling inter-variable dependencies using message passing \cite{pmlr-v70-gilmer17a}. The Adaptive Graph Convolutional Recurrent Network (AGCRN)  \cite{https://doi.org/10.48550/arxiv.2007.02842} is one of these models, that uses message passing to model inter-variable relations. It incorporates a Node Adaptive Parameter Learning (NAPL) module to capture node-specific patterns and a Data Adaptive Graph Generation (DAGG) module to infer inter-dependencies among different time series automatically. AGCRN outperforms state-of-the-art models on real-world spatio-temporal datasets without pre-defined graphs and spatial connections. 
Another model of this group is MtGNN \cite{https://doi.org/10.48550/arxiv.2008.08617}. It represents a notable advancement in forecasting models by incorporating graph propagation techniques. This approach allows the integration of neighborhood information from different hops, enabling the learning of high-order relations and substructures within the graph. MtGNN is designed to capture complex spatial and temporal dependencies in time series data. It also enhances temporal convolution
by utilizing multiple kernel sizes. 

To date, only a limited number of existing methods fall into the category of continuous models. In the realm of factorized methods, STGODE \cite{Fang_2021} proposes a depiction of graph propagation as a continuous process using a neural ordinary differential equation (NODE) \cite{NEURIPS2018_69386f6b}. This approach enables effective characterization of long-range spatial-temporal dependencies, incorporating dilated convolutions along the time axis. On the other hand, for coupled methods, MTGODE (Multivariate Time series with Graph neural Ordinary Differential Equations) \cite{Jin_2023} generalizes both spatial and temporal modeling processes found in most related works into a single unified process, integrating two NODEs. The MTGODE model was introduced to address key limitations in multivariate time series forecasting. It mitigates three primary challenges posed by existing methods. Firstly, in contrast to discrete neural architectures that interlace individually parameterized spatial and temporal blocks, resulting in discontinuous latent state trajectories and increased forecasting errors, MTGODE presents a continuous model. Secondly, it tackles high complexity by avoiding dedicated designs and redundant parameters found in discrete approaches, thereby reducing computational and memory overheads. Thirdly, it departs from reliance on predefined static graph structures, allowing for dynamic graphs with time-evolving node features and unknown graph structures.
MTGODE leverages neural differential equation techniques to complement missing graph topologies and unify spatial and temporal information transfer. Through deeper graph propagation and finely tuned temporal information aggregation, it characterizes stable and precise latent spatial-temporal dynamics. Experimental evaluations on five benchmark time series datasets demonstrate the superior performance of the MTGODE model across various metrics. 

The Graph-Coupled Oscillator Networks (GraphCON) \cite{rusch2022graphcoupled} framework introduces a novel approach to deep learning on graphs. It is built upon discretizations of a second-order system of ordinary differential equations (ODEs), representing a network of nonlinear controlled and damped oscillators coupled via the adjacency structure of the underlying graph. The framework exhibits flexibility by allowing any basic Graph Neural Network (GNN) layer to serve as the coupling function. Through this, a multi-layer deep neural network is constructed via the dynamics of the proposed ODEs. GraphCON addresses the oversmoothing problem often encountered in GNNs by relating it to the stability of steady states in the underlying ODE. The paper assumes that zero-Dirichlet energy steady states are not stable for the proposed ODEs, showcasing the ability of GraphCON to mitigate oversmoothing. Furthermore, the authors demonstrate, that GraphCON addresses the exploding and vanishing gradients problem, facilitating the training of deep multi-layer GNNs. As the GraphCON approach is substantiated through competitive performance compared to state-of-the-art methods across various graph-based learning tasks, it is used as wrapper for benchmarking in this paper.

By choosing the benchmarking models according to these four groups, we can evaluate their performance and compare it on a broader perspective. This comparison helps assess the strengths and weaknesses of different modeling techniques for anomaly detection, as they have varying abilities to capture different aspects of graph data.

\subsection{Additional Benchmarking Models}
In the last part of the results section, we also offer a comparison against other state-of-the-art deep geometric models for time-series and the MLP baseline in terms of anomaly detection. These models include MLP, TCNAE, VAE, TGCN and STGCN. 

The Multilayer Perceptron (MLP) \cite{gardner1998artificial} is a type of feedforward neural network characterized by its multiple layers of interconnected neurons. In an MLP, information flows through the network in one direction, from the input layer to the output layer. Each neuron in a layer is connected to every neuron in the subsequent layer, and each connection has an associated weight. The MLP uses activation functions to introduce non-linearity, allowing it to model complex relationships in data. MLPs are widely employed for various tasks, including classification and regression, owing to their capability to learn intricate patterns and representations from input data.

The TCNAE (Temporal Convolutional Network Autoencoder) model \cite{THILL2021107751} is an autoencoder variant designed to capture temporal dependencies in data. It employs a structure of temporal convolutional layers to extract temporal patterns effectively. The autoencoder comprises an encoder that reduces input data to a compact representation and a decoder that aims to reconstruct the original data from this representation. With a convolutional neural network (CNN)-like architecture, TCNAE adeptly captures both spatial and temporal dependencies in sequential data, such as time series or sensor data. The use of convolutional layers enables the model to learn localized patterns while considering the temporal structure of the data. TCNAE finds applications across domains, including signal processing, time series analysis, and tasks where the temporal dimension of the data is critical.

The Variational Autoencoder (VAE) \cite{Kingma_2019} is a generative model based on the autoencoder framework. Unlike traditional autoencoders, VAE incorporates probabilistic elements to model uncertain representations. The encoder transforms input data into a probability distribution in the latent space, determined by mean and standard deviation. The decoder then generates data points from random samples of this distribution. By introducing sampling in the latent space, VAE allows for the generation of more diverse and stochastic outputs compared to traditional autoencoders. VAE is applied in generative modeling, particularly for generating new data points, images, or other complex structures, while accounting for uncertain representations.

The Temporal Graph Convolutional Network (TGCN) \cite{Zhao_2020} is a specialized neural network designed for handling temporal graph-structured data. It extends traditional graph convolutional networks (GCNs) to incorporate temporal dynamics, making it suitable for time-evolving graph datasets. TGCN leverages both spatial and temporal information by applying graph convolutions across both dimensions. This enables the model to capture evolving patterns and dependencies in dynamic graph structures, making it particularly effective for tasks involving time-varying relationships. TGCN finds applications in various domains, including social network analysis, traffic forecasting, and any scenario where understanding temporal changes in graph-structured data is essential.

The Spatio-Temporal Graph Convolutional Network (STGCN) \cite{Yu_2018} is a specialized neural network designed to handle spatio-temporal data with a graph structure. It extends traditional graph convolutional networks (GCNs) to incorporate both spatial and temporal dependencies simultaneously. STGCN is particularly well-suited for tasks involving data that varies across both spatial and temporal dimensions, such as traffic flow prediction, weather forecasting, or video analysis. By leveraging graph convolutions in conjunction with temporal convolutions, STGCN captures complex spatio-temporal patterns, making it effective for understanding and predicting dynamic behaviors in graph-structured data over time.

\subsection{Physical Background}

The field of civil engineering and structural health monitoring has witnessed significant advancements, particularly in the detection and assessment of structural damage. This section reviews key contributions in civil engineering and structural health monitoring, focusing on the ModeConv approach.

Vibration-based techniques have been extensively explored as non-destructive methods for assessing structural integrity. Dai et al. \cite{dai2007free} categorized vibrations into free and forced types, where free vibrations occur when a structure oscillates on its own after displacement, and forced vibrations result from external energy application \cite{siedziako2017enhanced}. Modal analysis has been pivotal in studying vibration characteristics \cite{peeters2001stochastic}. Salawu \cite{salawu1995bridge} emphasized the importance of forced vibration testing, while ambient vibration testing utilizes natural vibrations during operation \cite{magalhaes2010damping}.

The ModeConv approach integrates deep learning with structural knowledge to analyze structural vibrations and detect changes. Statistical and spectral analysis techniques, including the Wiener-Chintschin relation \cite{hapel1990zufallsschwingungen}, PSD, and covariance, play crucial roles in this integration.

The Wiener-Chintschin relation \cite{hapel1990zufallsschwingungen} establishes a mathematical connection between Power Spectral Density (PSD) and the covariance-based autocorrelation function of a stochastic process. This relation enables ModeConv to analyze power spectral density, providing insights into the distribution of power across vibration frequencies.

PSD estimation \cite{hapel1990zufallsschwingungen} allows ModeConv to estimate the transfer function and coherence between the measured response and assumed excitation, essential for understanding structural behavior. Covariance \cite{stein2005space} computation forms the basis of the Signal Block, capturing similarity patterns between signals measured by accelerometers.

Incorporating the physical dynamics of structures is essential for accurate structural health monitoring. The equation of motion, relating structural motion to mass, stiffness, and damping, is fundamental \cite{peeters2001stochastic}. The PDE Block in ModeConv considers this equation, which is analyzed in the frequency domain using the Fourier transform \cite{cochran1967fast}. This analysis yields the frequency response function, crucial for understanding structural behavior.

The transfer function, derived from the frequency response function through the Laplace transform \cite{doetsch2013handbuch}, is crucial for ModeConv. It represents the relationship between input excitation and structural response, aiding in the identification of changes or anomalies.

By incorporating the equation of motion, frequency response function, and transfer function, the PDE Block of ModeConv integrates physical dynamics into the deep learning framework. The PDE Block provides a deeper understanding of the structure's response to external forces and vibrations.

\section{Datasets}
In this section, we introduce two significant datasets for structural health monitoring (SHM). The first dataset is the \emph{Simulated Smart Bridge dataset}, designed to offer controlled and diverse damage scenarios with varying severity levels. This artificial dataset enables comprehensive training and testing of machine learning algorithms, providing insights into the effects of different types and degrees of damage on sensor readings and network behavior.

The second dataset, the \emph{Luxemburg dataset}, is a real-world dataset that presents realistic damage scenarios observed in actual bridge structures. It incorporates environmental and operational factors, such as temperature and traffic loads, adding layers of complexity that are challenging to replicate in artificial datasets. Analyzing this dataset allows us to understand the behavior of sensor networks in practical situations and validate our methods with authentic, real-world data.

By leveraging both datasets, we aim to address the limitations inherent in each. The artificial dataset may lack the intricacies and uncertainties of real-world scenarios, while the real-world dataset might have fewer damage scenarios under controlled laboratory conditions. The integration of these datasets contributes to a broader understanding of machine learning models in structural health monitoring compared to many existing papers in the domain. The challenge for both datasets is the accurate distinction between external excitations and free vibrations, crucial for precise structural behavior analysis and anomaly detection. 

\subsection{Simulated Smart Bridge Dataset}

The Simulated Smart Bridge dataset was released by the Federal Highway Research Institute (BASt) \cite{freundt2020-09-30kuenstliche-messdaten} within the framework of the \textit{Intelligente Brücke} (smart bridge) initiative, a component of the \textit{Testfeld Digitale Autobahn} (digital highway test site) program under the auspices of the German Federal Ministry for Digital and Transport. The subject of observation in the smart bridge framework is a concrete highway bridge situated at a highway junction near Nuremberg, at the convergence of highway A3 from Frankfurt am Main to Regensburg and highway A9 from Ingolstadt to Bayreuth. The bridge spans four lanes, measuring a total length of 155 meters, and is supported by five concrete pillars, exhibiting a width of 15 meters.

During its construction in 2016, the eastern terminus of the bridge was outfitted with a wireless sensor network situated between the penultimate and final support columns. This network comprises 36 sensors strategically positioned across five distinct measurement cross-sections (see Fig. 2), measuring accelerations, elongations, and displacements of various bridge components. Given the recent construction of the bridge and its early stage in the operational lifecycle, the recorded data is deemed representative of the undamaged, healthy state.

\begin{figure}
    \centering
    \includegraphics[width=\textwidth]{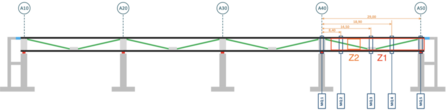}
    \caption{Side view of the measurement cross sections \cite{freundt2020-09-30kuenstliche-messdaten}.}
    \label{fig:pbim_bridge}
\end{figure}

In practical terms, the lack of real-world damage data required resorting to Finite Element Method simulations to simulate the baseline behavior of the bridge. The term ``baseline behavior'' here refers to the fundamental performance or condition of the bridge under normal conditions. This modeled behavior serves as a reference or baseline for subsequent comparisons with actual observations or measurements, especially when it comes to identifying damages or anomalies and was calibrated step by step according to the real measurements. Additionally, simulations incorporating three distinct damage scenarios (N, S1, S2 and S3) were conducted in an additional simulation project \cite{freundt2020-09-30kuenstliche-messdaten}. An overview of these scenarios is presented in Table \ref{tab:pbim_scenarios}. This dataset consists of artificial measurement data generated through a numerical simulation of a bridge with simplified geometry, allowing for controlled experiments and the comparison of different deep learning models for structural health monitoring. 

\begin{table}
    \caption{Overview of different damage scenarios in the Simulated Smart Bridge Dataset \cite{freundt2020-09-30kuenstliche-messdaten}.}
    \centering
    \begin{tabular}{ccl}
        \toprule
        Scenario & Duration (2018) & Description \\
        \midrule
        N  & 01.01, 00:00 -- 08.07, 23:59 & Healthy state\\
        S1 & 09.07, 00:00 -- 06.08, 23:59 & Gradual reduction of stiffness of $\unit[30]{\%}$ in Z1\\
        S2 & 09.07, 00:00 -- 06.08, 23:59 & Gradual reduction of stiffness of $\unit[50]{\%}$ in Z2\\
        S3 & 09.07, 00:00 -- 06.08, 23:59 & Immediate increase in the bearing stiffness\\
        \bottomrule
    \end{tabular}
    \label{tab:pbim_scenarios}
\end{table}

The tabular overview of the sensortypes and locations is presented in Table \ref{tab:pbim_sensors}. The strain gauges, strategically placed at different locations, measured the strain of the structure at a sampling rate of 75 Hz. Simulated environmental conditions included wind gusts modeled with a Gaussian stochastic process (mean wind speed of 30 m/s), traffic loads simulated through a moving force applied to the bridge, and temperatures ranging from -10°C to 40°C. The inclination sensors provide inclination values in milliradians. 

\begin{table}
    \centering
    \caption{Available sensors in the Simulated Smart Bridge dataset \cite{freundt2020-09-30kuenstliche-messdaten}.}
    \begin{tabular}{lcll}
        \toprule
        Sensor label & Section & Location & Sensortype \\
        \midrule
        MS\_U\_Neig & MQ1 & below & Inclination Sensor \\
        MS\_U\_Schieb & MQ1 & below & Inclination Sensor \\
        \midrule
        MS\_U\_MI\_L\_o & MQ2 & top & Strain Gauge \\
        Ms\_U\_Li\_Int\_u & MQ2 & bottom left & Strain Gauge \\
        MS\_O\_MI\_L\_u & MQ2 & bottom & Strain gauge \\
        MS\_U\_Neig & MQ2 & below & Inclination Sensor \\
        Ms\_U\_Re\_Int\_u & MQ2 & bottom right & Strain Gauge \\
        \midrule
        -- & MQ3 & \multicolumn{2}{c}{Same sensor setup as MQ2} \\
        \midrule
        -- & MQ4 & \multicolumn{2}{c}{Same sensor setup as MQ2} \\
        \midrule
        -- & MQ5 & \multicolumn{2}{c}{Same sensor setup as MQ1} \\
        \bottomrule
    \end{tabular}
    \label{tab:pbim_sensors}
\end{table}

\begin{figure}
    \tikzset{sensor/.style={draw, circle, fill=red, inner sep=1mm}}
    \tikzset{label/.style={draw, rectangle, font = \small}}
    \centering
    \begin{tikzpicture}
        \draw[fill=lightgray] (0,0) -- (4,0) -- (5,1) -- (9,1.5) -- (-5, 1.5) -- (-1,1) -- (0,0);
        \draw[fill=white] (0,0.5) -- (1.5,0.5) -- (1.7,0.35) -- (2.3,0.35) -- (2.5,0.5) -- (4,0.5)
              -- (4.7, 1.1) -- (4, 1.1) -- (3.8, 0.95) -- (0.2, 0.95) -- (0,1.1) -- (-.7,1.1) -- (0,0.5);
        \node[sensor] (bl) at (0.2,0.2) {};
        \path (bl) ++(-2,-0.5) node[label] {Floor left};
        \node[sensor] (below) at (2,-0.05) {};
        \path (below) ++(0,-0.8) node[label] {Below};
        \node[sensor] (floor) at (2,0.35) {};
        \path (floor) ++(-4,0.3) node[label] (floorLabel) {Floor};
        \draw [->, dashed] (floorLabel.east) -- (floor.north|-floorLabel.east) -- (floor.north);
        \node[sensor] (br) at (3.8 ,0.2) {};
        \path (br) ++(+2,-0.5) node[label] {Floor right};
        \node[sensor] (top) at (2,1.1) {};
        \path (top) ++(0,1) node[label] {Top};
    \end{tikzpicture}
    \caption{Measurement cross-section of the simulated bridge structure \cite{freundt2020-09-30kuenstliche-messdaten}.}
    \label{fig:pbim_sensor_positions}
\end{figure}
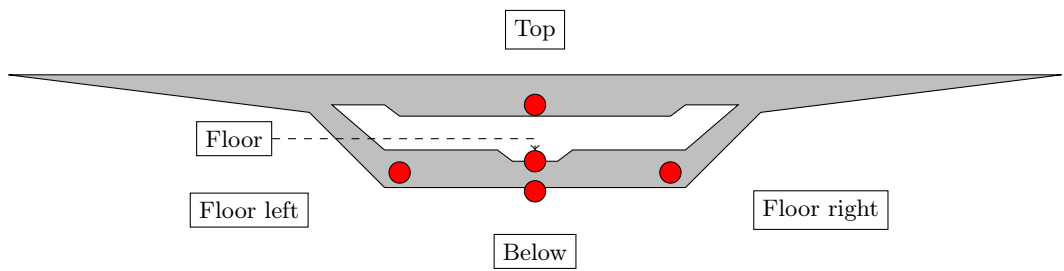

The positions of the sensors within the cross sections are indicated in Fig. \ref{fig:pbim_sensor_positions}. Specifically, the sensors have beenplaced either inside the hollow chamber on the floor at the center, on the floor to the right, on the floor to the left, below the concrete of the hollow chamber, or on top the concrete of the hollow chamber. Placing sensors at evenly distributed points across the hollow chamber ensures comprehensive coverage of the area under surveillance, minimizing the risk of overlooking important information.\\

\subsection{Luxemburg Dataset}
The Luxemburg dataset, as documented in Schommer et al. \cite{schommer2017damage}, comprises structural vibration measurements obtained from a decommissioned bridge that underwent various real-world damage scenarios. This bridge, originally constructed from 1953 to 1955, spanned the River Mosel, connecting the German city of Wellen to the city of Grevenmacher in Luxemburg.

The bridge consisted of five spans, each composed of five parallel concrete beams. Steel tendons interconnected neighboring beams, with additional transverse concrete beams placed every $\unit7.5{m}$ along their longitudinal axis. Each beam was pre-stressed by tendons embedded into the concrete along its longitudinal axis.
The sensor network contained three different types of sensors: 26 accelerometers, eight displacement sensors, voltage sensors and temperature sensors. All displacement sensors were mounted to the bottom of the beam with their reference fixed to the ground below. Seven of these measured vertical displacement while SH7 measured horizontal displacement. The sensor network employed three types of sensors: 26 accelerometers, eight displacement sensors, voltage sensors, and temperature sensors. Displacement sensors were affixed to the bottom of the beam, with their reference fixed to the ground. Seven sensors measured vertical displacement, while one, denoted as SH7, measured horizontal displacement \cite{schommer2017damage}. Temperature sensors were embedded into the concrete at a depth of at least $\unit5{cm}$, measuring the ambient temperature in the shadow. To simulate the original bridge's weight, weights totaling $\unit30{t}$ were permanently placed on the left half of the beam. 

To simulate the weight of the original bridge, several weights with a total mass of $\unit30{t}$ were permanently placed on the left half of the beam. 

\begin{figure}[htbp]%
 \centering
 \includegraphics[width=1.1\textwidth]{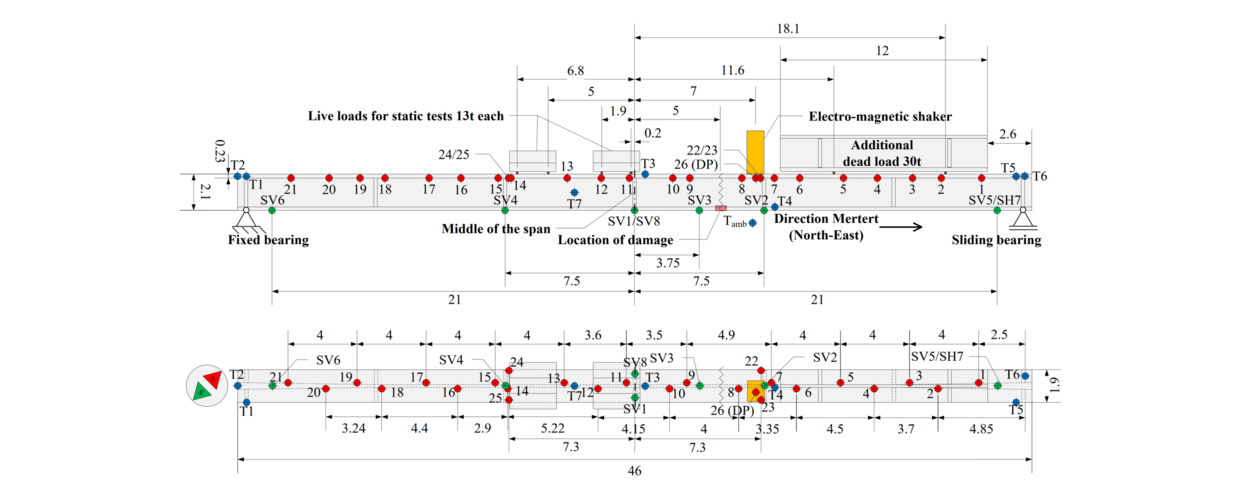}
 \caption{Overview of the measurement setup in side view (upper view) and top viewer (lower part) with a, accelerometers in red; b, displacement sensors in green; c, temperature sensors in blue and d, shakers in yellow, \cite{schommer2017damage}.}\label{Querschnitt}
 \end{figure}

The dataset encompasses both static and dynamic tests. Static testing involved temporarily placing up to four live loads of $\unit13{t}$ each on the bridge. Dynamic tests assessed the vibration response under external excitation using a shaker (see Fig. \ref{Querschnitt}) , applying a swept sine wave with a sweep rate of $\unitfrac[0.02]{Hz}{s}$ and a range from $\unit2.5{Hz}$ to $\unit10{Hz}$ with an excitation force of $\unit2000{N}$ \cite{schommer2017damage}.

The averaging of temperature values from seven sensors is done to represent an overall mean temperature for the structure. This approach is grounded in the principle of spatial averaging, where multiple measurements are combined to obtain an average or representative value. By utilizing several sensors, local variations or irregularities in the temperature distribution can be mitigated, and the averaged temperature provides a more stable and reliable representation of the overall structural temperature. This is particularly crucial when the structure is subject to varying environmental influences in different areas. 

Acceleration data, recorded at a sampling rate of $\unit2500{Hz}$, is parsed from TDMS-files in the \texttt{DynamicMeasurements} folder using the npTDMS Python library\footnote{See \url{https://nptdms.readthedocs.io/en/stable/}}. Temperature measurements, occurring once per minute, are associated with the nearest acceleration measurement.

Following the bridge's demolition, two beams were transported to the port of Mertert in Luxemburg for experiments. The beams, each $\unit46{m}$ in length and weighing approximately $\unit120{t}$, underwent various damage scenarios involving cutting tendons and adding masses at specific locations. These scenarios, outlined in Table \ref{tab:lux_scenarios}, encompass progressive damage stages. Sensor network details are illustrated in Fig. \ref{Querschnitt}.

\begin{table}
    \centering
    \caption[Overview over the different damage scenarios in the Luxemburg dataset]{Overview over the different damage scenarios in the Luxemburg Bridge Dataset \cite{schommer2017damage}. The damage is applied progressively, i.e. the changes made in scenario S1 are still present in all following scenarios.}
    \begin{tabular}{ccl}
        \toprule
         Scenario & Duration (2014) & Description \\
         \midrule
         N & 23.01, 00:00 -- 31.01, 12:00 & Healthy state \\
         S1 & 31.01, 12:00 -- 04.02, 12:00 & 2 tendons cut \\
         S2 & 04.02, 12:00 -- 06.02, 12:00 & 4 tendons cut \\
         S3 & 06.02, 12:00 -- 11.02, 12:00 & 6 tendons cut \\
         S4 & 11.02, 12:00 -- 19.02, 12:00 & 6 tendons cut, 6 tendons partly cut\\
         \bottomrule
    \end{tabular}
    \label{tab:lux_scenarios}
\end{table}

\subsection{Data Preprocessing}
The total size of both datasets in their raw unzipped format amounts to 2.3TB for the simulated dataset and 68.7GB for the Luxemburg dataset. Besides the experiments conducted on these large datasets, that can be requested from the university of Luxembourg and from the German Federal Highway Research Institute (Bundesanstalt für Straßenwesen) in full size, we stored five percent of the datasets in csv format in the provided open repository\footnote{\url{https://github.com/MilanShao/ModeConv}}. We didn't exclude any data of the datasets.
In the case of the Luxemburg dataset, the signals have been recorded by a data acquisition system (DAQ) of National Instruments and stored in TDMS data format, while the smart bridge simulated data was delivered in a binary data format. As the initial format of both datasets was not suitable for time-series analysis, we devised a preprocessing pipeline to transform the data into an intermediate format. For each day of measurements, we generate a separate file for each measurement channel of all sensors with the original sampling rates and resolutions.  The data is stored, with each measurement encoded as its timestamp (4-byte integer) followed by the measured value at that time (4-byte float). Finally, we split the data into disjoint training, validation, and test sets as follows and normalize them. Over the course of 23 (non-consecutive) days and 52 experiments, data were collected for the normal behavior class (10 \%) and the abnormal behavior class (90 \%) containing all four damage classes. For training, experiments with IDs $[0, 1, 2, 3, 6]$ were used, while experiments with IDs $[5, 7, 22, 51]$ were allocated for validation. The remaining experiments were designated for testing. Experiments $[0:7]$ were characterized by normal behavior, while the rest were labeled as exhibiting at least one of the damage classes.

In the case of the Simulated Smart Bridge dataset data were collected over a course of 217 days towards the classes $N$ (189 days of normal behavior) and $S0$-$S3$ (28 days split between normal and three distinct damage scenarios representing abnormal behavior). This results in a distribution of approximately 80\% normal behavior and 20\% abnormal behavior. In the provided five percent dataset all classes have been used equally.

\section{ModeConv Method}
\begin{figure}[htbp]%
\centering
\includegraphics[width=1.0\textwidth]{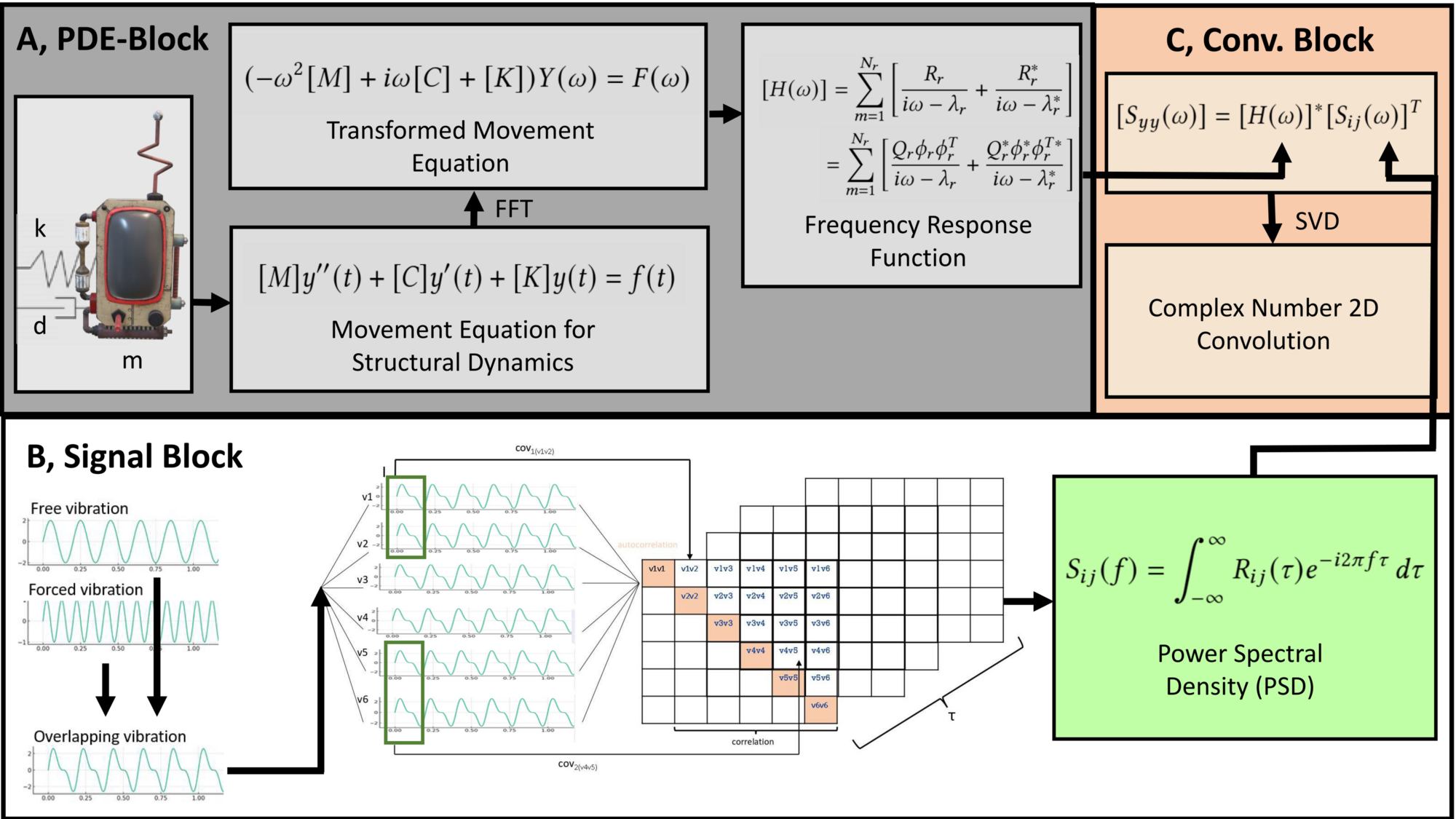}
\caption{Overview over ModeConv building blocks with its A, ODE Block, B, Signal Block and C, Convolutional Block.}\label{Modeconfoverview}
\end{figure}

Vibration-based techniques are employed to assess the integrity of structures. We categorize vibrations into three types: free vibrations, which occur when a structure oscillates on its own, and forced vibrations, caused by external energy sources like wind gusts or traffic-induced vibrations ~\cite{dai2007free, siedziako2017enhanced} as well as their overlapping vibrations (see Fig. \ref{Modeconfoverview}, Signal Block). By extracting eigenmodes, thus focusing on free vibrations of the structure, we can detect changes and damages in the structure ~\cite{peeters2001stochastic, salawu1995bridge}. We therefore calculate the frequency response function of the PDE building block (see Fig. \ref{Modeconfoverview}, PDE building block) as well as the Power Spectral Density (PSD) (see Fig. \ref{Modeconfoverview}, Signal Block) as input to the Convolutional Block (see Fig. \ref{Modeconfoverview}, Conv. Block). To enable anomaly detection in structures, ModeConv leverages statistical and spectral analysis techniques to analyze the vibration behavior. Specifically, it utilizes the modal transformation of the signals contained in the datasets to identify any deviations or anomalies in the structural response.

The method section is organized into the three aforementioned main building blocks of the overview figure: the PDE building block, the signal block, and the convolution block, that are depicted in the ModeConv Overview (see Fig. \ref{Modeconfoverview}). Each of these building blocks is explained in detail in the following subsections. Please note that whenever a calculation of one entry in a matrix is described, the notation is written without square brackets, while matrices for which every entry is calculated contain square brackets. All matrices are written with capital letters, while vectors are written with small letters, and tensors in bold letters.

\subsection{Data Input}
\begin{figure}[htbp]%
\centering
\includegraphics[trim=0cm 1cm 0cm 1cm, clip, width=1.0\textwidth]{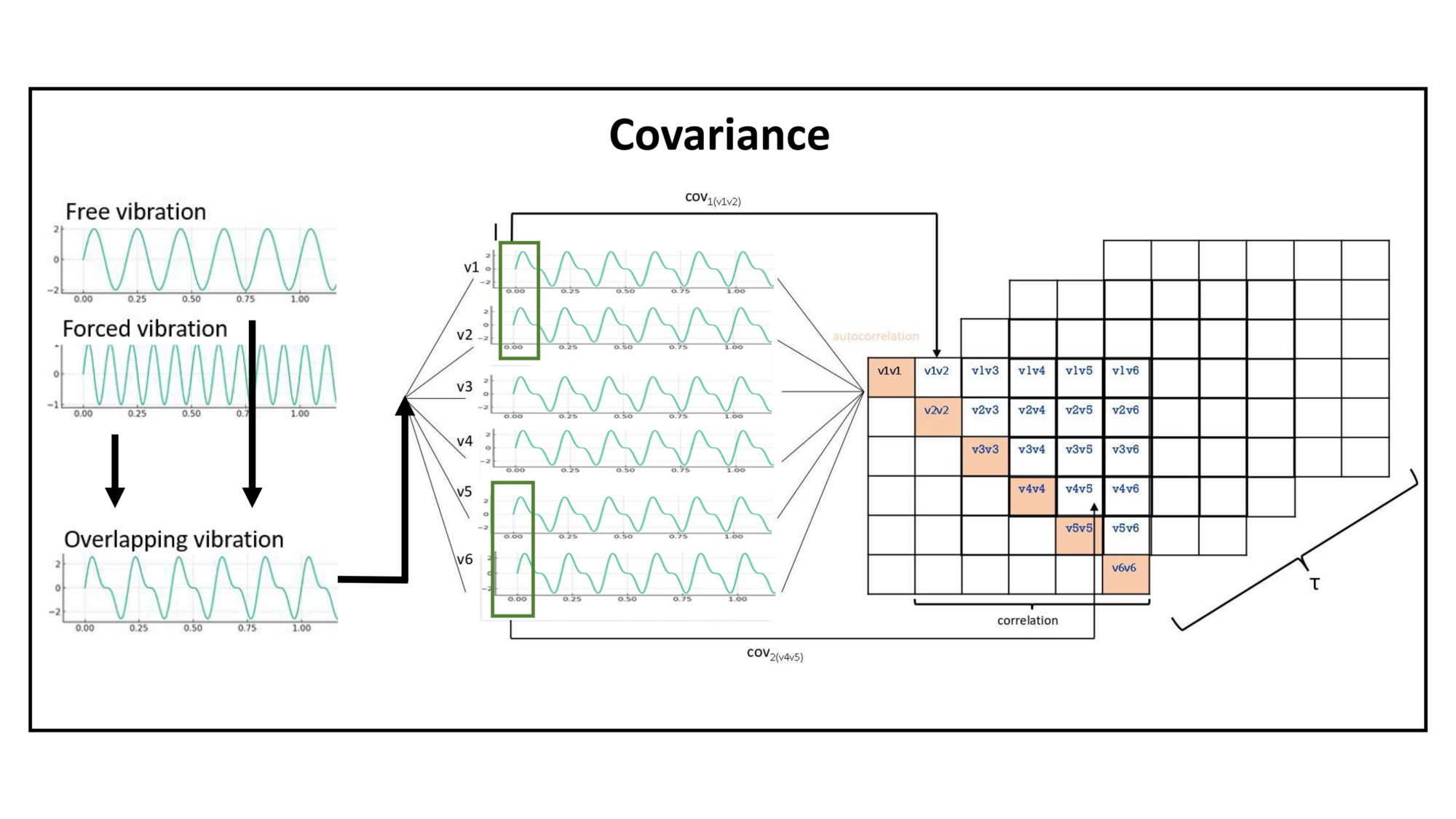}
\caption{Overview of ModeConv Covariance: Overlapping signals are inserted as input for all multivariate time-series. The covariance of signal pairs is then calculated over five samples for every correlation in the covariance matrix. Subsequently, the time delay $\tau$ over the covariance matrices is determined.}\label{Covariance}
\end{figure}

In applications of sensor networks, data naturally reside on the vertices and edges of weighted graphs, in which the sensors represent the vertices and the connection between the sensor nodes can be represented by edges. This graph structure can be used in structure aware models such as graph convolutions.

A Graph is represented as $G = (V,E)$ where $V$ is the set of vertices ($v_i$) representing sensor nodes and $E$ is the set of edges between sensor nodes. The data on these graphs is represented as a finite set of samples, with each sample, known as a sensor measurement $x_i(t)$ for a sensor node $v_i$, associated with each vertex in the graph. The edge index is denoted as $e = (v_i, v_j)$ where $0 \leq i,j < |v|.$\\
Additional dataset-specific sensor readings can be introduced as vertex features in this graph model, depending on the dataset. In the Luxemburg dataset, it includes measurements of acceleration, temperature, voltage, and displacement. For voltage, temperature, and strain features, statistics such as mean, standard deviation, minimum, and maximum values are computed. In the smart bridge simulated damage dataset, we obtain strain-gauge, temperature, and acceleration measurements. One sample is obtained by concatenating the mean values of strain, temperature, and acceleration along the last dimension. The batch-size has been optimized towards 256.

Additionally, acceleration measurements are specifically used for pairwise signal covariance calculation, denoted as $C_{ij}$ (illustrated as signals in Fig. \ref{Covariance}). Considering the number of sensors $N$ in the network, a single sample is constructed by concatenating the mean values of voltage, displacement, and temperature along the last dimension.

\subsection{Signal Block}
\begin{figure}[htbp]%
\centering
\includegraphics[trim=0cm 2cm 0cm 2cm, clip, width=1.0\textwidth]{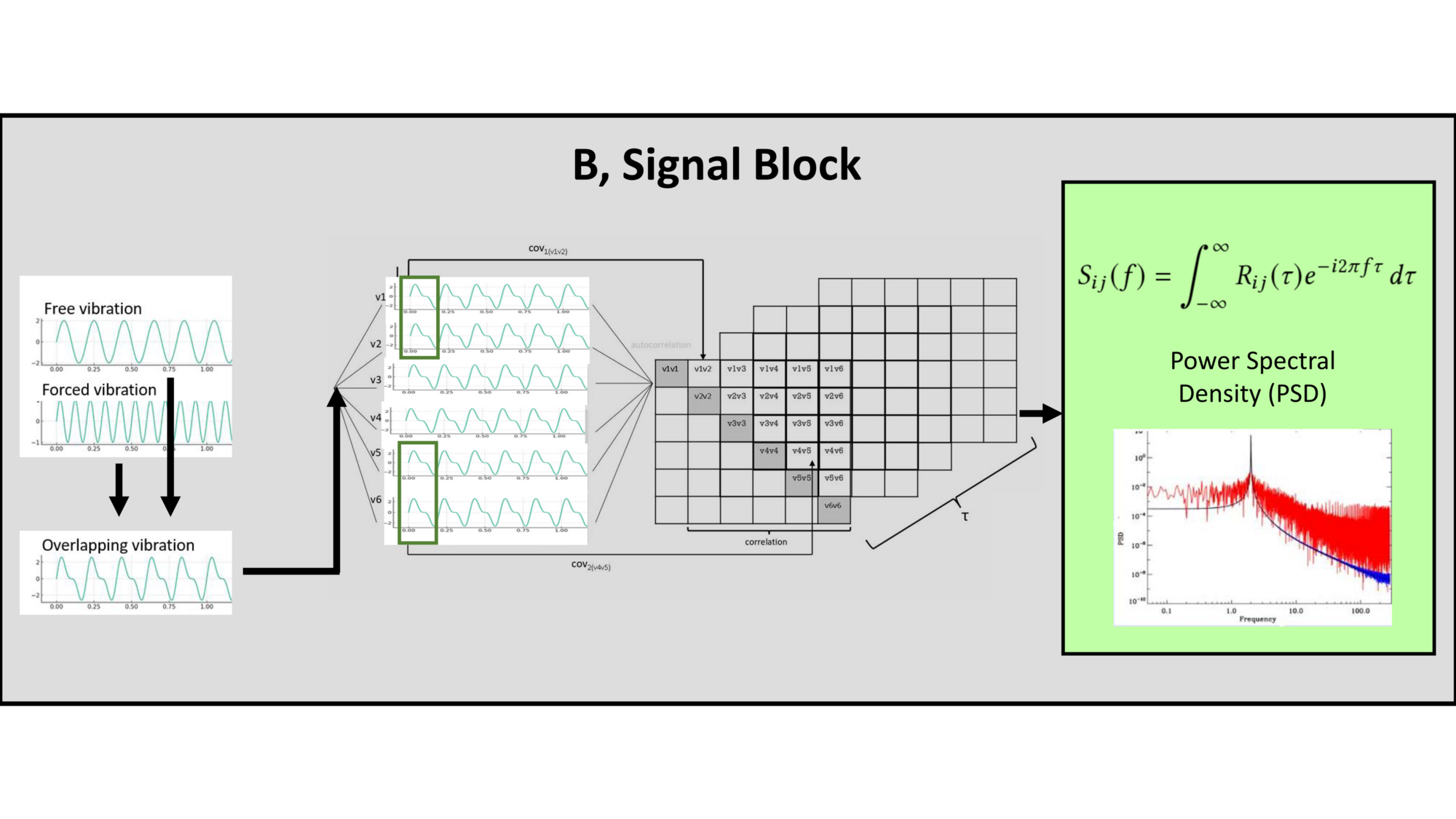}
\caption{Overview over ModeConv signal block: the covariance of signals is taken as input to calculare the Power Spectral Density (PSD).}\label{Signalblock}
\end{figure}
In the context of signal processing, a crucial component lies within the signal block, where specific strategies are employed to process and analyze the data streams effectively. Within this signal block only the first sensor channel is used for the calculations, because these are the relevant data to calculate the eigenmodes while additional data like temperature show dependecies and are therefore just used as additional features. In the case of the Luxemburg dataset that is the acceleration measurement and in the case of the simulated dataset the strain gauges are used (see section above).\\ 

To ensure that the time series data can be evenly divided into overlapping windows of size $l$, the data is first padded. The hyperparameter $l$ has been optimized for five samples, which corresponds to an aggregation interval of microseconds given a sampling rate of 2500 samples per second. Larger aggregation steps were found to result in decreased performance, despite the desire to reduce the size of the large datasets. Therefore $l$ was kept small. Then the covariance $C_{ij}$ for pairwise signals is computed for all signals $x_i$ and $x_j$, where $i$ and $j$ is used to seperate the measurements of the sensors $i$ and $j$, over the length of $l$ samples. In this equation, $x_i$ and $x_j$ are two different signals $i$ and $j$  and $x_i(t)$ is the value of the signal $x_i$ at time-step $t$ for a sensor node $v_i$.

$$X_i = \{ x_i(l) : 0 \leq l < 5 \}$$
$$X_j(\tau) = \{ x_j(l + \tau) : 0 \leq l < 5 \}$$
$$C_{ij}(\tau) = {E}[(X_i - {E}[X_i])-(X_j(\tau) - {E}[X_j(\tau)])]$$
\begin{equation}
 C_{ij}(\tau) = {E}[(x_i(l) - \mu_i)(x_j(l+\tau) - \mu_j)]  
\end{equation}

The transformation from the cross-correlation $C_{ij}(\tau)$ to the cross-correlation $R_{ij}(\tau)$ is achieved through scaling. The cross-correlation $R_{ij}(\tau)$ is defined as the normalized version of the covariance function $C_{ij}(\tau)$.

To transition from $C_{ij}(\tau)$ to $R_{ij}(\tau)$, we divide $C_{ij}(\tau)$ by the square root of the product of the variances of signals $x_i$ and $x_j$~\cite{magalhaes2011explaining}:

\begin{equation}
R_{ij}(\tau) = \frac{C_{ij}(\tau)}{\sqrt{C_{ii}(0)C_{jj}(0)}}
\end{equation}

Here, $C_{ii}(0)$ represents the autocovariance of signal $x_i$, and $C_{jj}(0)$ represents the autocovariance of signal $x_j$ at zero time lag ($\tau = 0$). This normalization yields the normalized cross-correlation function $R_{ij}(\tau)$, which is commonly used for further analysis.

The Fourier transformation of the cross-correlation $R_{ij}(\tau)$ yields the power spectral density (PSD) $S_{ij}(f)$, describing the distribution of power across frequencies in the frequency domain.\\ 

\begin{equation}
{S}_{ij}(f) = \int_{-\infty}^{\infty} R_{ij}(\tau) e^{-i2\pi f \tau} \, d\tau
\label{eq:sij}
\end{equation}

This relation is called the Wiener-Chintschin relation, which is a mathematical relationship that allows to analyze the power spectral density (PSD) of structural response signals. This provides valuable information about the distribution of power across different vibration frequencies~\cite{hapel1990zufallsschwingungen}.

Calculating covariances, cross-correlations, and power spectral density (PSD) in the Signal Block provides deeper insights into the relationships and structures within the sensor data. This allows to recognize patterns, identify trends, and gain crucial insights into the behavior of the system. PSD offers information about the distribution of power across various frequencies. Covariance and cross-correlation reveal the relationships between different signals or sensors, facilitating the identification of common patterns in the data. Analyzing PSD and cross-correlation enables the detection of changes in system behavior over time, facilitating early detection of anomalies indicating potential problems or deviations in the system.

\subsection{PDE Building Block}

\begin{figure}[htbp]%
\centering
\includegraphics[trim=0cm 2cm 0cm 2cm, clip, width=1.0\textwidth]{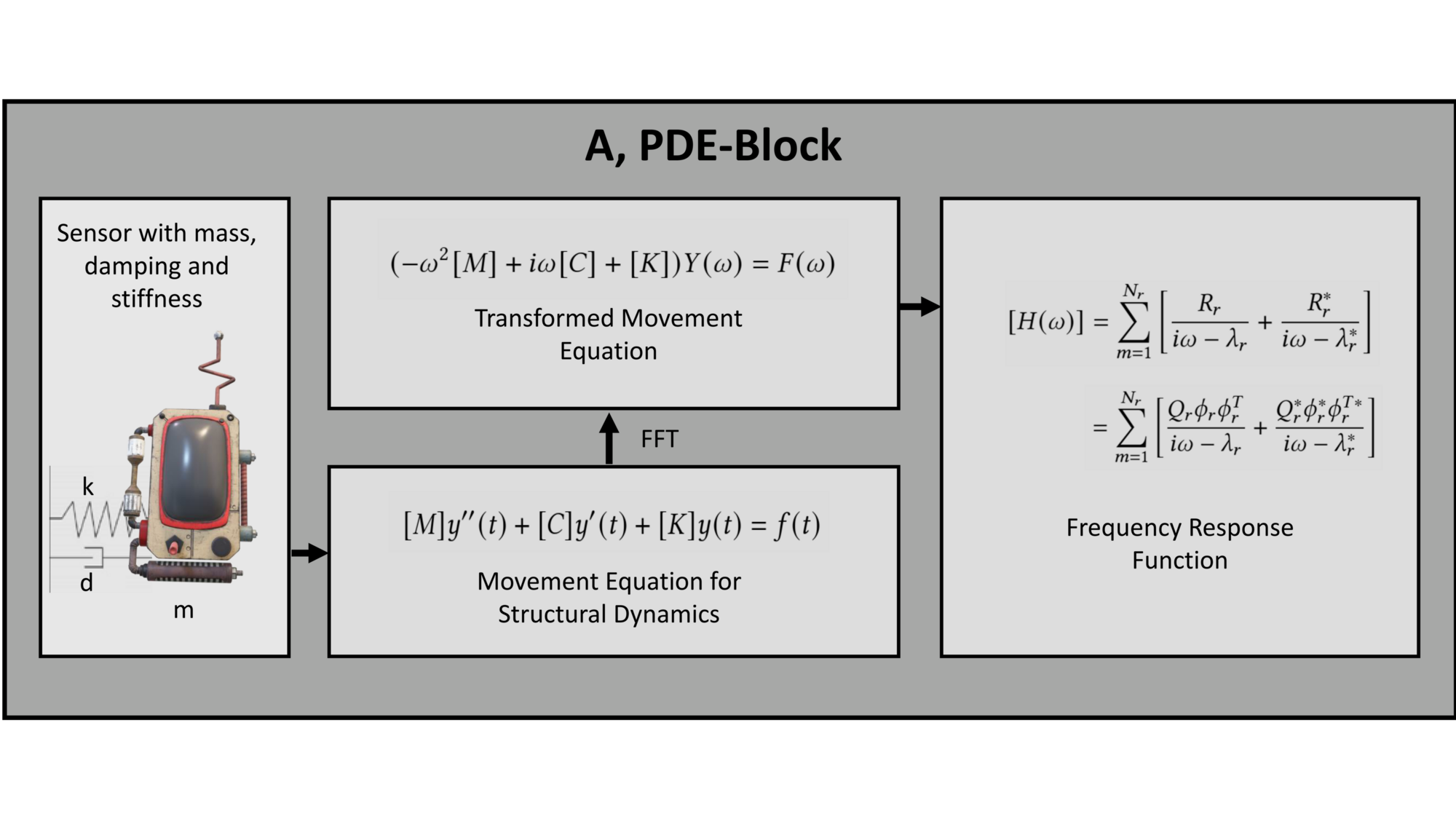}
\caption{Overview over ModeConv PDE block with the vector of excitation forces $f(t)$, $m$ denoting the mass point of each sensor, $[M]$ denoting the mass matrix, $d$ denoting the damping, $[C]$ denoting the damping matrix, $k$ denoting the stiffness and $[K]$ denoting the stiffness matrix. $y''(t)$ denotes the vector of acceleration, $y'(t)$ the vector of velocity and $y(t)$ the vector of displacement in the movement equation for structural dynamics. In the second step the Fast Fourier Transformation (FFT) is calculated on this movement equation to gain the transformed result. The transformed result is then used to calculate the frequency response function.}\label{PDE}
\end{figure}

The equation of motion describes the dynamics of a system and enables the determination of its natural frequencies, damping ratios, and modes. By modeling the equation of motion of a system, it is possible to develop a mathematical understanding of the system behavior. This is particularly useful for analyzing the structural or material properties of constructions or machinery, such as bridges, buildings, aircraft, or turbines.

The PDE building block (see Fig. \ref{PDE}) makes use of this motivation and expresses the dynamic behavior of a system in the time domain using the equation of motion \cite{Dinkler2020}. In the figure, $d$ denotes damping, $m$ denotes mass and $k$ denotes stiffness. The equation of motion for structures includes a vector of excitation forces, denoted as $f(t)$. In both datasets, the mass of the structure or infrastructure is divided into points of mass according to the number of sensors, and this sums up in a mass matrix $[M]$. The mass matrix $[M]$ is used for analyzing the behavior of structures subjected to dynamic loads such as earthquakes, wind, or traffic. It is a diagonal matrix with the mass of the structure, proportionally allocated to each sensor mass point denoted by $m$ as one mass point on the diagonal, i.e. the degrees of freedom are equal to the number of sensor nodes. In this context, each node $|v|$ of the graph possesses its individual mass denoted by $m$. $y''(t)$ denotes the vector of acceleration, $y'(t)$ the vector of velocity and $y(t)$ the vector of displacement. The dimensions of the stiffness matrix $[K]$ of the structure are equal to the number of sensors in the network, and as a bridge can be defined as a one-mass-swinger it follows a series connection. In this context, a series connection refers to the configuration where the bridge is considered as a sequence of interconnected elements. Here, the number of sensors in the network corresponds to the dimensions of the stiffness matrix $[K]$, implying that each sensor serves as a component in this series arrangement. The damping of the system is expressed through the damping matrix of the structure $[C]$:

\begin{equation}
[M]y''(t) + [C]y'(t) + [K]y(t) = f(t)
\label{eq:eqmotion}
\end{equation}

The stiffness matrix $[K]$ is defined in series connection as~\cite{huang2002duality}:

\begin{equation}
[K]_{(i,j)} =
\begin{cases}
    2k & \text{if } i = j \\
    -k & \text{if } \lvert i - j \rvert = 1 \\
    0 & \text{otherwise}
\end{cases}
\end{equation}

The dimensions of the stiffness matrix $[K]$ are also equal to the number of sensors in the network.

In order to decouple the equation of motion \cref{eq:eqmotion}, it is transformed to the modal form~\cite{patil2000decoupled}. The basic idea is to express every oscillation state of a system by a combination of its natural modes. This is achieved by transforming the time-dependent displacements into a modal coordinate frame~\cite{Dinkler2020}
\begin{equation}
x(t) = [\phi]p(t) 
\end{equation}
where $[\phi]$ denotes the modal matrix and $p(t)$ the modal coordinates.

Then the eigenvalue problem is solved ~\cite{peeters2001stochastic}. This returns the natural frequencies $f_n$ and mode shapes $\phi$. The mode shapes are then normalized, so that the maximum value of each column is equal to 1. Next, the mode shapes are ordered by ascending natural frequency. 

The modal mass matrix [$M_{modal}$] is calculated by projecting the mass matrix onto the normalized mode shapes\cite{Dinkler2020}. The same is done for the modal stiffness matrix [$K_{modal}$]
\begin{equation}
    [M_{modal}] = [\phi]^T\Lambda[\phi]
\end{equation}
\begin{equation}
    [K_{modal}] = [\phi]^T\Gamma[\phi],
\end{equation}

Here, the damping ratio $\xi$ is set to 0.02 for concrete material. The modal damping matrix [$C_{modal}$] is obtained by transforming [$C_{modal}$] into the coordinate system of the original mass matrix using the inverse of the mode shape matrix $[\phi]^{-1}$.

The mode shape matrix $[\phi]$ represents the spatial patterns of vibrations associated with the natural frequencies of the system. By utilizing these mode shapes, the modal mass, stiffness, and damping can be expressed in a coordinate-system-friendly framework, facilitating the analysis and interpretation of structural properties.

To diagonalize the damping matrix [$C$] with the eigenvectors of the undamped system, the convenience hypothesis \cite{de1982damped} must be introduced. In this case, proportional damping is assumed. In the special case of Rayleigh damping  \cite{liu1995formulation}, [$C$] can be expressed as a linear combination of the mass and stiffness matrices.

Then the equation of motion in frequency space can be denoted as:

\begin{equation}
(-\omega^2[M] + i\omega[C] + [K])Y(\omega) = F(\omega)
\end{equation}

Here, $Y(\omega)$ represents the response of the system in frequency domain, while $F(\omega)$ represents the external excitation force. $[M]$, $[C]$, and $[K]$ are the mass, damping, and stiffness matrices of the system. We follow the calculation of Rainieri et al.~\cite{rainieri2014operational}

The matrix of system responses $[H(\omega)]$ is expressed as a sum of resonance responses. Using partial fraction decomposition $[H(\omega)]$ can be expressed in its modal form ~\cite{rainieri2014operational}:

\begin{equation}
    [H(\omega)] = \frac{Y (\omega)} {F(\omega)} = \frac{\text{adj}(-\omega^2[M] + i\omega[C] + [K])}{\text{det}(-\omega^2[M] + i\omega[C] + [K])}
\end{equation}

The matrix of system responses $[H(\omega)]$ can be denoted as ~\cite{rainieri2014operational}:

\begin{equation}
    [H(\omega)] = \sum_{r=1}^{N_r} \frac{[R_r]}{i\omega - \lambda_r} + \frac{[R_r^*]}{i\omega - \lambda_r^*} \\
    = \sum_{r=1}^{N_r} \frac{Q_r\phi_r\phi_r^T}{i\omega - \lambda_r} + \frac{Q_r^*\phi_r^*\phi_r^{T*}}{i\omega - \lambda_r^*}
\end{equation}\\

Here $N_r$ represents the number of mode shapes of the system, while $\lambda_r = \sigma_r + i\omega_{d,r}$ is denoted as the pole of the $r$-th mode and $\phi_r$ is the mode shape of the $r$-th eigenmode, while $Q_r$ denotes the scaling factor. The scaling factor $Q_r$ is introduced to modulate the contribution of the $r$-th mode to the overall system response. Specifically, $Q_r$ is multiplied with the outer product of the mode shape $\phi_r$ in the expression for the matrix of system responses $[H(\omega)]$. The role of $Q_r$ is to control the amplitude or intensity of the $r$-th mode within the total system response. Its introduction allows for the adjustment of the influence of each mode on the dynamic behavior of the system. Please note that the $*$ is used for the complex conjugate, $\lambda_r$ represents the conjugate pole element of the $r$-th mode, $\phi_r$ denotes the conjugate mode shape element of the $r$-th eigenmode, and $Q_r^*$ signifies the complex conjugate scaling factor of the $r$-th mode. $R_r$ represents the contribution of the $r-th$ mode in the frequency domain. It denotes the residue term of the partial fraction decomposition of the frequency response function.

\subsection{Convolution Block}

\begin{figure}[htbp]%
\centering
\includegraphics[width=1.0\textwidth]{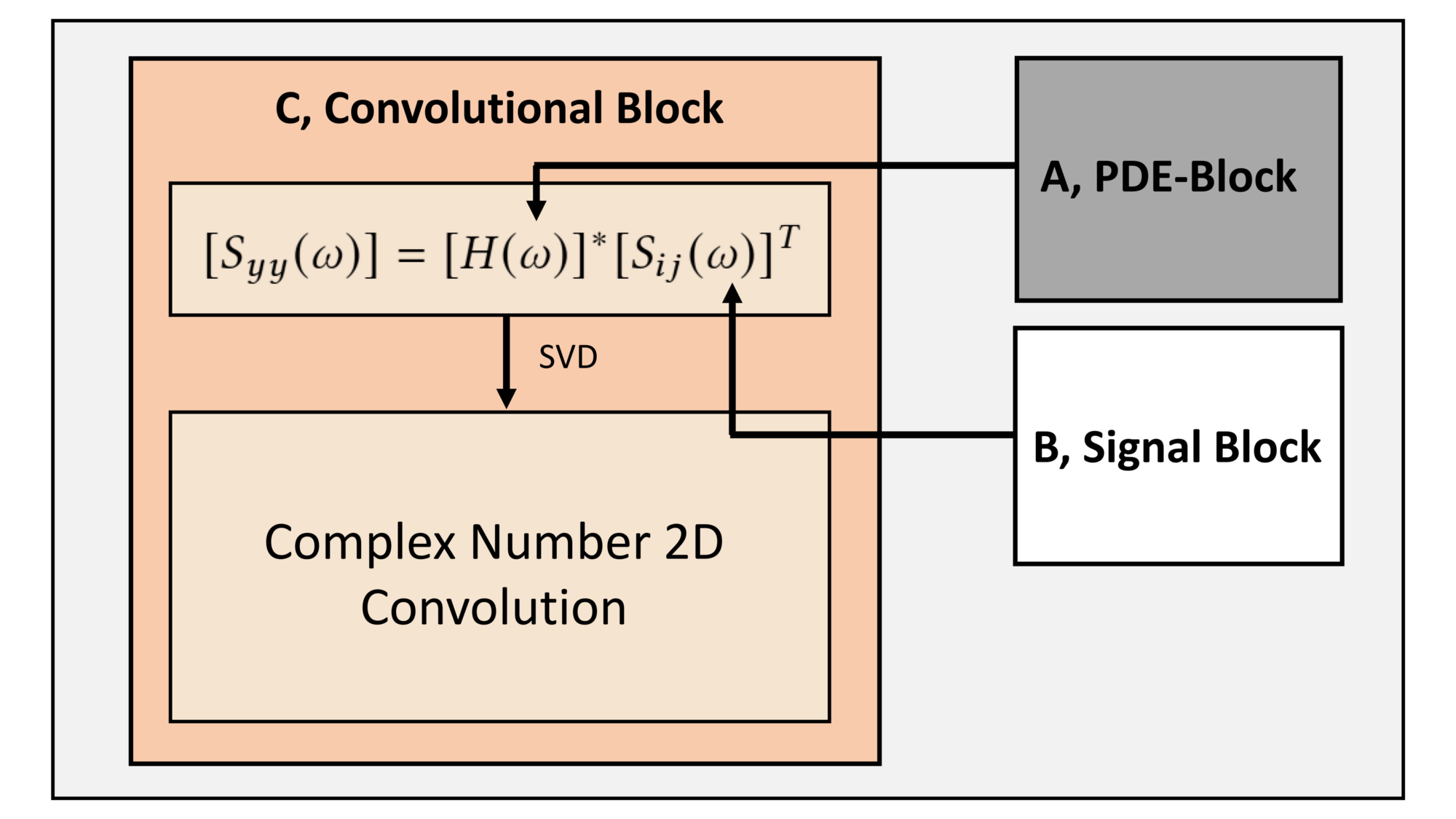}
\caption{Overview over ModeConv Convolutional block: the results of the frequency response function from the PDE block as well as the Power Spectral Density calculation of the signal block are taken as input of the Convolutional block to calculate the $S_{yy}(\omega)$. Then the Singular Value Decomposition (SVD) is calculated and taken as input to the convolutional filter with two dimensions, where ${y^\sim}$ is the output signal, ${W}_r$ and $\mathbf{W}_i$ are the weight matrices for the real and imaginary components and ${x}_r$ and ${x}_{im}$ denote the real and imaginary components of the input signal.}\label{Convblock}
\end{figure}

In the Convolution Block, the Power Spectral Density (PSD) matrix [$S_{ij}$] in its diagonal form is obtained from the Signal Block (see Fig. \ref{Convblock}). It is element-wise multiplied with the complex-valued weighting matrix [${H(\omega)}$] to obtain the weighted PSD matrix $[S_{yy}] \in \mathbb{C}$. It is computed as follows~\cite{rainieri2014operational}:

\begin{equation}
[S_{yy}(\omega)] = [S_{ij}(\omega)][H(\omega)]^T
\end{equation}

Here, $S_{ij}$ denotes the cross-power spectral density matrix as given in \cref{eq:sij}\cite{rainieri2014operational}. In the literature, there can be found different versions of multiplications for these matrices~\cite{magalhaes2011explaining, rainieri2014operational}. It is important to note, that the cross-power spectral density matrix is typically symmetric. Therefore, transposing $[S_{ij}(\omega)]$ yields the same information than $[S_{ij}(\omega)]$. Therefore, this results in the weighting matrix when multiplied with $[H(\omega)]^T$. 

\paragraph{Convolutional Filter}
In Graph Signal Processing, the goal of the convolutional filter is to capture the underlying structure or patterns within the graph data. When dealing with modal coordinates, that are uncorrelated, $[S_{yy}(\omega)]$ shows a diagonal structure, wherein all values outside of the diagonal are zero. We then perform a singular value decomposition (SVD) on $[S_{yy}(\omega)]$ for specific frequencies per batch. This allows $[S_{yy}(\omega)]$ to be factorized and can be expressed as: 

\begin{equation}
    [S_{yy}(\omega)] = [U][\mathlarger{\varepsilon}][V]^T
\end{equation}

Here, $\text{diag}[\mathlarger{\varepsilon}]$ contains the singular values in descending order, while $[U]$ and $[V]$ contain the singular vectors. Direct comparison with modal decomposition shows that the individual singular values of the SVD can be regarded as a measure of the contribution of each mode shape to the overall system response. The singular vectors represent the mode shapes of the system. Therefore, the system can be decomposed into as many single-degree-of-freedom oscillators as the system has degrees of freedom, and their interaction can be expressed through the modal coordinates in the form of $[\mathlarger{\varepsilon}]$. 

After the singular value decomposition, the matrix $[U]$ is utilized to compute the filter weights for the convolutional filter. These weights indicate the contribution of each eigenmode to the overall structure or dynamics of the processed data. Subsequently, these weights are passed to the graph neural network as part of the ModeConv layer. %
 
They determine how information from neighboring nodes is propagated and aggregated through the network. By capturing and processing structural information based on the material properties of infrastructures, these physics-informed weights can be regarded as essential part of the ModeConv layer and are incorporated into the ModeConv Laplace variant as follows.
In the propagate function, %
each node's features are combined with those of its neighbors, ensuring that the output retains the same shape as the input. 
Within each GNN layer, message passing is utilized to exchange and aggregate information over neighboring nodes by graph convolution. %
The filter matrix is complex, with two components representing the real and imaginary parts. 
As the weights $W$ are also from the complex domain, their real part ${W}_r$ and imaginary part ${W}_i$ are both incorporated as weights into the complex-valued message passing scheme reflecting the significance of the messages between neighboring node.
In contrast, the filter design in ModeConv Fast is  defined as follows.
In the ModeConv Fast filter, both the input tensor and the eigenmodes tensor resulting from the SVD operation are used. %
The filter is computed as the tensor product of these two tensors. %

In summary, integrating engineering-specific filter weights aids to capture and encode domain-specific knowledge. Thus, the model is enabled to learn the modal representations automatically and is therefore supported to perform tasks like anomaly detection, classification or regression on graph-structured data, that are based on changes in material properties. %
The final output of the model is obtained by stacking the output $y$ from each layer.

\subsection{Anomaly detection with ModeConv layers}

The ModeConv layer can then be used in an Graph Autoencoder setting to detect anomalies based on changes in the eigenmodes. Besides the anomaly detection task, one could also use ModeConv for classification or regression tasks, although these tasks are not performed in the experiments.

In the case of anomaly detection, the models are conditioned on an (semi-)supervised surrogate task such as reconstructing a given output through a bottleneck
or predicting the evolution of an input in the future. During inference, anomalies can then be detected via a threshold mechanism on the reconstruction/prediction errors or residuals of the model. Given a multi-variate time-series with signals $x_{ij} \in \mathbb{R}$, the model $M$ is asked to predict the multi-variate time-series $Y \in \mathbb{R}$ immediately following $x_{ij}$. Its prediction $\hat{Y}$ is then compared to the ground-truth $Y$ via a loss function $L$. Here, the MSE between $Y$ and $\hat{Y}$ are used.

Anomaly detection in this context is carried out within the framework of (semi-)supervised learning, where the model is trained based on the given data of the normal class of both datasets. ModeConv is taken as layer of a GNN Autoencoder. The model is trained to reconstruct the input data. During training, the model learns to reproduce the input data as accurately as possible. When the model is applied to data that does not conform to the training patterns, reconstruction errors can serve as indicators of anomalies. A higher reconstruction error indicates a greater deviation from normal patterns. Once the model is trained, thresholds are set to determine at what point a reconstruction error is considered an anomaly.

Here, the effectiveness of two different methods for setting the threshold in anomaly detection tasks is assessed: the L1 threshold \cite{malkauthekar2013analysis} and the Mahalanobis distance \cite{de2000mahalanobis} calculated at the 95th percentile. These measures are employed to detect anomalies in various models. The L1 threshold method involves determining a threshold value based on the L1 norm. The L1 norm represents the sum of the absolute values of the elements in a vector. In this context, the L1 norm is calculated for each data point, and a threshold is set. Any data point with an L1 norm exceeding this threshold is considered an anomaly. The Mahalanobis distance is calculated at the 95th percentile. This distance provides a measure of how far a data point deviates from the mean in multivariate space, considering the covariance structure of the data. Points beyond the 95th percentile threshold are identified as anomalies. The Mahalanobis distance is particularly suitable for situations with a larger variance within the data, as it accounts for correlations between different features. In the anomaly detection process, each data point is evaluated based on the chosen method (L1 threshold or Mahalanobis distance). If the calculated metric for a data point exceeds the set threshold, the point is flagged as an anomaly. This process enables the identification of unusual patterns or outliers within the dataset.

On the other hand, the Mahalanobis distance is a measure of the distance between a data point and the mean of a distribution, taking into account the covariance matrix. We calculate the Mahalanobis distance for each data point and compare it against the Mahalanobis distance at the 95th percentile. If the Mahalanobis distance of a data point exceeds the 95th percentile value, it is identified as an anomaly.

\section{Experiments}

We propose two major advantages of ModeConv, which need to be validated in the experimental section. First, ModeConv delivers better results on anomaly detection tasks that rely on physical material properties and their monitoring compared to existing graph neural network models for multivariate time-series. Second, ModeConv reduces the data to the modal coordinates and uses SVD instead of the normalized Laplacian while retaining relevant information, resulting in better performance in terms of runtime.

To validate these claims, we compare the results of a graph autoencoder with ModeConv layers trained on the Luxembourg dataset with the results of the same graph autoencoder using ChebConv layers \cite{defferrard2017convolutional}, which are a type of Spectral Graph Convolution. We also evaluate the results on the Simulated Smart Bridge dataset. In addition, we compare the performance of ModeConv autoencoder for the whole datasets with two state-of-the-art Graph Neural Networks for multivariate time-series, namely AGCRN \cite{https://doi.org/10.48550/arxiv.2007.02842} and MtGNN \cite{wu2020comprehensive}, both of which are built on Spectral Graph Convolutions \cite{cvetkovic1980spectra}. For the five percent dataset we also compare it against the MTGODE model and the GraphCON wrapper.

All of the models have been tuned via a hyperparameter study using optuna \cite{akiba2019optuna}, that aims to improve the reconstruction error for all networks in order to create a fair experimental setup. The parameters included in the paper have been learning rate, chebyshev filter size $K$ from ${2,...,8}$, number of layers ranging from ${1, 3, 5, 10}$, the number of hidden dimensions ranging from ${4, 8, 16}$ and the bottleneck ranging from ${1, 2, 4}$. Learning rate and dropout have been optimized according to the default values of Pytorch Lightning.

For the GraphCON model \cite{rusch2022graphcoupled} the following formalization is used as foundation for the wrapper: \[X'' = \sigma(F\Theta(X, t)) - \gamma X - \alpha X',\] \\
where \(\sigma\) is the activation function, \(F\Theta\) is the 1-neighbourhood coupling function, and \(\alpha, \gamma\) are learnable hyperparameters providing dampening. \(X\) denotes the time-dependent matrix of node features over time $t$, which serves as the main input for the forward function between GNN layers. In the case of GraphCON, we therefore also tune the learnable hyperparameters \[0 \leq \alpha, \gamma \leq 2.\] \\

In their GitHub repository\footnote{\url{https://github.com/tk-rusch/GraphCON}}, Rusch et al. provide a code snippet, that had to be modified for the experiments to use it as wrapper for the MtGNN model. 

The existing MtGNN architecture relies on the trainer functionality provided by PyTorch Lightning, which necessitates the model to be a PyTorch Lightning module. This requirement is not met by GraphCON. Consequently, we incorporated a wrapper directly into the message passing of MtGNN. However, GraphCON specifically expects graph convolutional layers with stable dimensions as input.

The MtGNN model consists of alternating graph convolutions (GC) and temporal convolutions (TC). The TC includes Dilated Inception Layers that lead to dimension reduction. This posed a challenge as GraphCON requires stable dimensions for its input. To address this, we made additional adaptations. In order to ensure that the dimensions of \(X\) and \(Y\) remain equal across iterations, a TC module is added for \(Y\).

Moreover, we compare our ModeConv Fast version without normalized Laplacian with the ModeConv Laplace version  in all experiments. Every model has been trained for 50 epochs. All experiments have been conducted on an NVIDIA GeForce GTX 1080 Ti GPU.

Furthermore, we compare the training time of all the models and conduct a comparative study on the single blocks of the ModeConv approach. This study evaluates the time-consuming and cost-improving aspects of the layer to accurately measure the complexity and performance improvement.

\section{Results}
To compare the results, several standard performance measures including precision, recall, F1 score, balanced accuracy, AUC (Area Under the Curve), and training time per epoch have been used. Precision indicates the accuracy of anomaly detection, with higher values representing fewer false positives while recall measures the ability to detect actual anomalies, with higher values indicating better performance. F1 Score is a balanced measure of precision and recall, with higher scores indicating a better trade-off between the two. Furthermore we used balanced accuracy, that calculates the average accuracy for both positive and negative classes, considering both highly imbalanced datasets used in these experiments. Moreover, AUC (Area Under the Curve) represents the model's ability to distinguish between positive and negative samples \cite{irizarry2019introduction}. The time taken for each epoch during training, measured in hours has been tracked to document the energy costs.

\subsection{Results Comparative Study}
An comparative study was conducted on the Luxemburg Dataset to analyze the ModeConv layer's components and their impact on performance. Two variations were examined: replacing Power Spectral Density (PSD) calculation with two linear layers (LL) in the signal block, and using normalized Laplacian instead of Single Value Decomposition (SVD) of the Convolutional block for the complex filter based on Laplacian weights.

\begin{figure}[htbp]%
\centering
\includegraphics[width=1.0\textwidth]{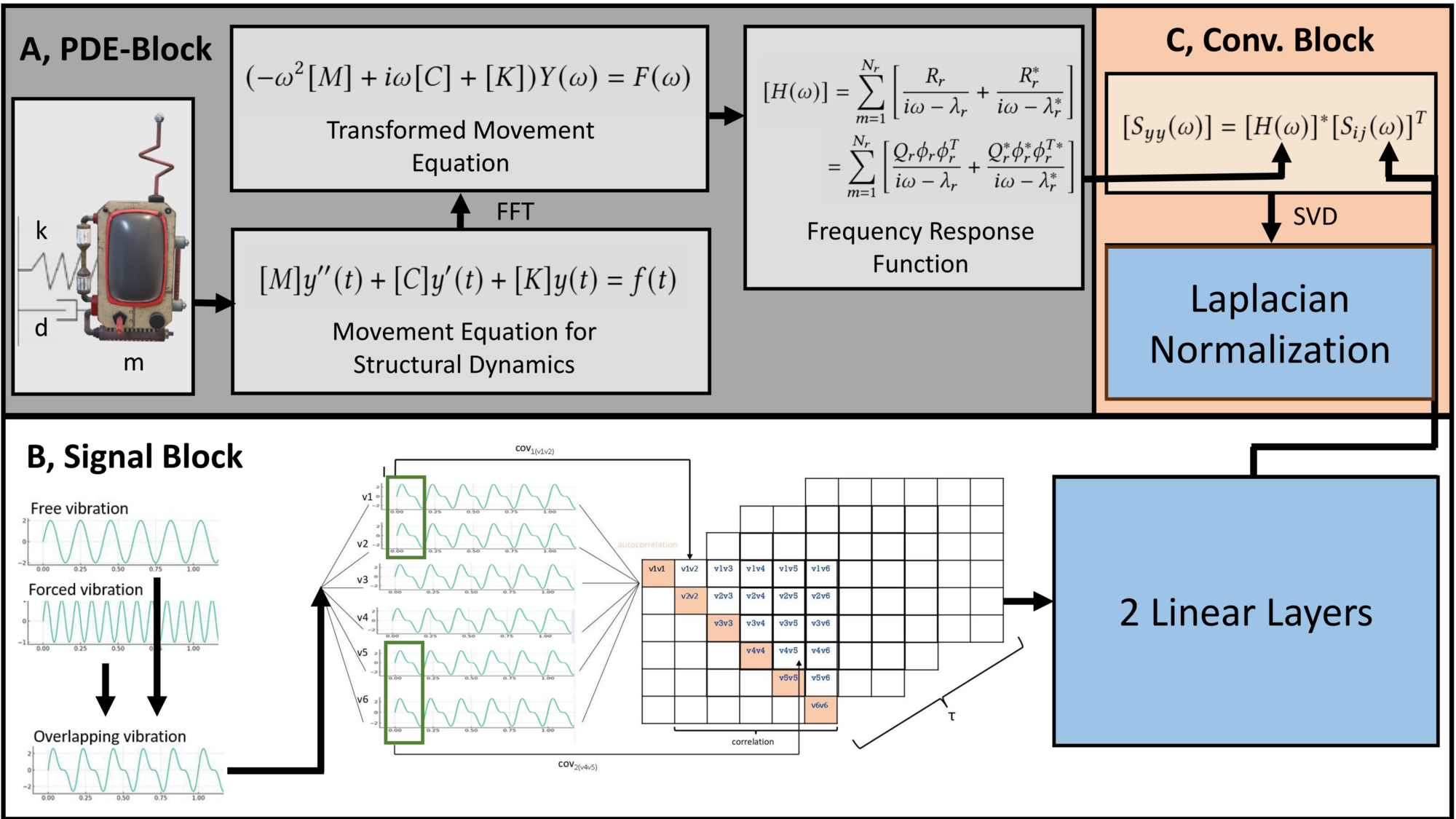}
\caption{ModeConv comparative study: instead of conducting the Singular Value Decomposition (SVD), the Laplacian Normalization is used like in ChebConv and instead of the Power Spectral Density (PSD) calculation, two linear layers are inserted as comparison.}\label{Ablation}
\end{figure}

\begin{table}[h]
\centering
\caption{Performance metrics for different variants of the ModeConv model performance on the Luxemburg dataset. The best results are highlighted in grey.}
\footnotesize
\renewcommand{\arraystretch}{1.2}
\setlength{\tabcolsep}{2pt}
\begin{tabular}{lrrrrr}
\toprule 
Model Variants & AUC & Bal. Acc. & F1 & Prec & Recall \\
\midrule 
ModeConv with SVD and PSD & 0.91 & 0.87 & 0.90 & 0.98 & 0.84 \\
ModeConv with Laplacian & 0.91 & 0.87 & 0.90 & 0.98 & 0.84 \\
\rowcolor[HTML]{E6E6E6} ModeConv with LL and Lap. & \textbf{0.92} & \textbf{0.88} & \textbf{0.91} & \textbf{0.99} & \textbf{0.85} \\
\bottomrule
\end{tabular}
\label{tab:ModeConv-variants_1}
\end{table}

The performance of the different ModeConv variants was evaluated, and the results are summarized in \cref{tab:ModeConv-variants_1}. All variants achieved similar performance. While the comparative study shows comparable performance, the speed or computational efficiency of the ModeConv with SVD and PSD variant, which is used as ModeConvFast in the following experiments, is higher whenever $K$ is low (see following tables). This evaluation shows, that the modal transformation delivers comparable results to the spectral transformation.

\subsection{Results Luxemburg Dataset}
Regarding the whole Luxemburg dataset, the results for different models using the 95th percentile of L1 distance as the threshold are presented in \cref{tab:models}.
\begin{table}[h]
\centering
\caption{Performance metrics for different models on the Luxemburg dataset with L1 distance. Best results highlighted in grey.}
\footnotesize
\renewcommand{\arraystretch}{1.2}
\setlength{\tabcolsep}{2pt}
\begin{tabular}{@{}lrrrrrr@{}}
\toprule
Model & Prec. & Recall & F1 & Bal. Acc & AUC & Time (h/epoch) \\
\midrule
\rowcolor[HTML]{E6E6E6}
ModeConvFast & 1.0 & 0.903 & 0.949 & \textbf{0.952} & \textbf{0.985} & \textbf{2} \\
ModeConvLaplace & 0.99 & \textbf{0.904} & \textbf{0.949} & \textbf{0.952} & 0.984 & 19 \\
ChebConv & 1.0 & 0.902 & \textbf{0.950} & 0.948 & 0.983 & 19 \\
MtGNN & 1.0 & 0.900 & 0.947 & 0.950 & 0.976 & 15 \\
AGCRN  & 1.0 & 0.898 & 0.946 & 0.949 & 0.973 & 18 \\
\bottomrule
\end{tabular} 
\label{tab:models}
\end{table}
ModeConvFast outperformed other models, achieving the highest F1 score of 0.949, perfect precision, and a recall of 0.903. It also obtained the highest balanced accuracy of 0.952 and AUC of 0.985, with the lowest training time per epoch, while the training time per epoch with 2 hours versus the highest value between 15-19 hours was 7-9 times higher. As the Laplacian normalization is used in ChebConv as well as ModeConvLaplace the training times show off to be similar.

Furthermore, when using the Mahalanobis distance as the threshold, ModeConvFast achieved an F1 score of 0.939 and a recall score of 0.907, making it the best-performing model. ModeConvLaplace obtained the highest precision score of 0.994 and a balanced accuracy of 0.868, while ChebConv and AGCRN also showed competitive performance.

\begin{table}[h]
\centering
\caption{Performance metrics for different models on the Luxemburg dataset with Mahalanobis distance. Best results highlighted in grey.}
\footnotesize
\renewcommand{\arraystretch}{1.2}
\setlength{\tabcolsep}{2pt}
\begin{tabular}{lrrrrrr}
\toprule
Model & Prec. & Recall & F1 & Bal. Acc & AUC \\
\midrule
\rowcolor[HTML]{E6E6E6}
ModeConvFast & 0.974 & \textbf{0.907} & \textbf{0.939} & 0.665 & 0.911 \\
ModeConvLaplace & 0.994 & 0.865 & 0.925 & \textbf{0.868} & \textbf{0.935} \\
ChebConv & 0.993 & 0.554 & 0.711 & 0.735 & 0.765 \\
MtGNN & 0.500 & - & 0 & - & 0.500 \\
AGCRN  & \textbf{1.0} & 0.543 & 0.704 & 0.772 & 0.771 \\
\bottomrule
\end{tabular} 
\label{tab:lux_maha}
\end{table}

In summary, the choice of threshold metric has a notable impact on model performance. ModeConvFast demonstrated superior performance with L1 distance, while ModeConvLaplace excelled with Mahalanobis distance. The selection should be based on specific requirements, considering precision, recall, balanced accuracy, and training time trade-offs (Table \ref{tab:lux_maha}).

\subsection{Results Simulated Smart Bridge Dataset}

Regarding the whole Simulated Smart Bridge Dataset when using the L1 distance to choose the threshold, the following results are gained:

\begin{table}[h]
\centering
\caption{Performance metrics for different models on the Simulated Smart Bridge dataset with L1 distance. Best results highlighted in grey.}
\footnotesize
\renewcommand{\arraystretch}{1.2}
\setlength{\tabcolsep}{2pt}
\begin{tabular}{@{}lrrrrrr@{}}
\toprule
Model & Prec. & Recall & F1 & Bal. Acc & AUC & Time (h/epoch) \\
\midrule
\rowcolor[HTML]{E6E6E6}
ModeConvFast & \textbf{0.688} & 0.618 & \textbf{0.626} & 0.754 & 0.826 & \textbf{1}\\
ModeConvLaplace & 0.531 & \textbf{0.661} & 0.603 & \textbf{0.757} & \textbf{0.829} & 15\\
ChebConv & 0.271 & 0.615 & 0.376 & 0.601 & 0.636 & 14\\
MtGNN & 0.630 & 0.603 & 0.616 & \textbf{0.757} & 0.828 & 16\\
AGCRN  & 0.403 & 0.514 & 0.452 & 0.662 & 0.724 & 8\\
\bottomrule
\end{tabular} 
\label{tab:Sim_L1}
\end{table}

The ModeConvLaplace model achieved the highest balanced accuracy score of 0.757, indicating that it performed well on both the positive and negative classes. Its AUC score of 0.829 was also the highest among the models. 
The ModeConvFast model achieved the highest precision score of 0.688, which indicates that it had a low false positive rate. However, its recall score of 0.618 was lower than that of the ModeConvLaplace model, which achieved the highest recall score of 0.661. The ModeConvFast model also had the highest F1 score of 0.626, indicating that it had a good balance between precision and recall. The AUC score of the ModeConvFast model was 0.826, which was the second highest among the models. Again the training time per epoch is much faster than the runtime of all other models. 

When using the Mahalanobis distance instead we gain the following results:

\begin{table}[h]
\centering
\caption{Performance metrics for different models on the Simulated Smart Bridge dataset with Mahalanobis distance. Best results highlighted in grey.}
\footnotesize
\renewcommand{\arraystretch}{1.2}
\setlength{\tabcolsep}{2pt}
\begin{tabular}{lrrrrrr}
\toprule
Model & Prec. & Recall & F1 & Bal. Acc & AUC \\
\midrule
\rowcolor[HTML]{E6E6E6}
ModeConvFast & \textbf{0.997} & 0.923 & \textbf{0.977} & 0.949 & 0.978 \\
ModeConvLaplace & \textbf{0.997} & 0.958 & 0.962 & 0.960 & 0.975 \\
ChebConv & - & 0 & - & 0.5 & 0.5 \\
MtGNN & 0.895 & \textbf{0.968} & 0.930 & \textbf{0.970} & \textbf{0.996} \\
AGCRN  & 0.792 & 0.933 & 0.857 & 0.936 & 0.985 \\
\bottomrule
\end{tabular} 
\label{tab:sim_maha}
\end{table}

Among these models, ModeConvFast and ModeConvLaplace stand out with precision scores of 0.997, showcasing their ability to accurately identify anomalies. ModeConvFast achieves an impressive F1 score of 0.977, while ModeConvLaplace excels in recall with a value of 0.958, highlighting its capability to capture a high proportion of actual anomalies. MtGNN also demonstrates strong performance with a balanced accuracy of 0.970 and an AUC of 0.996. 

\subsection{Results Luxemburg Dataset 5\%}

For the experiments with the reduced dataset, we additionally compare the results against the MtGNN with GraphCON wrapper and against MTGODE in terms of F1 Score:

\begin{table}[ht]
\centering
\caption{F1 Scores for Different Models}
\begin{tabular}{l c}
\toprule
Model & F1 \\
\midrule
\rowcolor{gray!20} ModeConvFast & \textbf{93.09} \\
ModeConvLaplace & 73.74 \\
ChebConv & 74.53 \\
AGCRN & 86.67 \\
MtGNN & 82.00 \\
MtGNN with GraphCON & 92.20 \\
MTGODE & 66.00 \\
\bottomrule
\end{tabular}

\end{table}

ModeConvFast achieves the highest F1 score of 93.09, outperforming other models. ModeConvLaplace, ChebConv, and AGCRN exhibit F1 scores ranging from 73.74 to 86.67. MtGNN achieves an F1 score of 82.00, whereas the incorporation of the GraphCON wrapper improves its performance to 92.20. In contrast, MTGODE lags behind with an F1 score of 66.00. In comparison to the outstanding results of the MTGODE paper, the model performs less competitively on this dataset. One potential reason for this could be the presence of outliers and noise among the data samples, which were not included in the datasets used in the original MTGODE paper. It is likely that the training datasets in the original MTGODE paper did not contain as many outliers and noise as the datasets used in the Luxemburg and simulated smart bridge datasets. \\
What we also observe is a notable enhancement in the performance of ModeConv when computing on the reduced dataset compared to other models. This improvement accentuates its robustness, suggesting that the model excels in handling both small and large datasets. %
This versatility positions ModeConv as choice for anomaly detection tasks, showcasing its ability to maintain efficacy even with limited data. 

\subsection{Results Simulated Smart Bridge Dataset 5\%}

As the complete dataset is several terabytes in size and thus computationally demanding to process, we also conducted experiments on 5 percent of the  dataset. As well as for the reduced Luxemburg dataset, we also conducted experiments on the reduced simulated smart bridge dataset. Additionally, we integrated the GraphCON approach here, as it closely resembles ModeConv and therefore represented a relevant extension for our experiments.

\begin{table}[ht]
\centering
\caption{F1 Scores for different Models}
\begin{tabular}{l c}
\toprule
Model & F1 \\
\midrule
ModeConvFast & 87.93 \\
ModeConvLaplace & 88.07 \\
ChebConv & 83.89 \\
AGCRN & 87.76 \\
MtGNN & 86.78 \\
\rowcolor{gray!20} MtGNN with GraphCON & \textbf{90.67} \\
MTGODE & 63.00 \\
\bottomrule
\end{tabular}
\end{table}

The presented results underscore the efficacy of the GraphCON approach, which corresponds to the author's advice against simply stacking Graph Neural Networks (GNNs) \cite{rusch2022graphcoupled}. The result validate the potential for significant enhancements in existing GNN architectures through GraphCON.  While GraphCON introduces two additional learnable hyperparameters demanding careful optimization for optimal results, the computational load is the primary drawback. Although GraphCON's direct impact on training time is marginal, its true benefits emerge with increased model depth, amplifying the number of trainable parameters and subsequently extending training time. 
ModeConv Fast demonstrates 3 percent less effective results to the GraphCON wrapper while significantly reducing runtime. 

The following pseudocode outlines the forward pass of the GraphCON wrapper, which is designed to capture dynamic behaviors through the application of Ordinary Differential Equations (ODEs). The algorithm utilizes a series of Graph Neural Network (GNN) layers to iteratively update two variables, denoted as \(X\) and \(Y\). The update process is governed by an Implicit-Explicit (IMEX) scheme, providing a numerical solution to the ODEs. Furthermore, the algorithm supports optional dropout regularization during training. This pseudocode serves as a representation of the underlying PyTorch implementation, adapted for this specific setting.

\begin{algorithm}
\caption{GraphCON Forward Pass}\label{graphcon}
\begin{algorithmic}[1]
    \Procedure{GraphCON}{$GNNs, X0, Y0, \text{edge\_index}, dt, \alpha, \gamma, \text{dropout}$}
        \State $X \gets X0$
        \State $Y \gets Y0$
        
        \For{$gnn$ \textbf{in} $GNNs$}
            \State $Y \gets Y + dt \cdot (\text{ReLU}(gnn(X, \text{edge\_index})) - \alpha \cdot Y - \gamma \cdot X)$
            \State $X \gets X + dt \cdot Y$
        \EndFor
        
        \If{$\text{dropout} \neq \text{None}$}
            \State $Y \gets \text{Dropout}(Y, \text{dropout})$
            \State $X \gets \text{Dropout}(X, \text{dropout})$
        \EndIf
        
        \State \textbf{return} $X, Y$
    \EndProcedure
\end{algorithmic}
\end{algorithm}

\section{Results of the additional benchmarking models}

In this section different state of the art models for anomaly detection in time-series are used additionally and the benchmarking results are presented. The additional models contain MLP, VRAE, TCNAE, TGCN and STGCN. MtGNN, AGCRN and ModeConv are inserted in the chapters before.

The following table shows the results of ModeConv in comparison to these models for the Luxemburg dataset:

\begin{table}[h]
\centering
\caption{Results for the complete Luxemburg dataset: ModeConv with L1 distance against other state of the art models}
\begin{tabular}{lcccc}
\toprule
Model & Precision (\%) & Recall (\%) & F1-Score (\%) & Accuracy (\%) \\
\midrule
MLP & 74.71 & 10.91 & 30.74 & 68.30 \\
VRAE & 56.32 & 10.89 & 26.72 & 65.24 \\
TCNAE & 74.86 & 18.72 & 25.76 & 63.86 \\
TGCN & 77.24 & 1.09 & 21.32 & 55.29 \\
STGCN & 38.89 & 3.64 & 16.73 & 55.71 \\
MTGNN & 59.29 & 6.63 & 33.73 & 67.69 \\
AGCRN & 54.84 & 8.67 & 34.38 & 61.82 \\
ModeConvFast & \textbf{99.99} & \textbf{90.30} & \textbf{94.90} & \textbf{95.20} \\
\bottomrule
\end{tabular}
\label{tab:my_table}
\end{table}

In analyzing further results on the Luxemburg dataset, ModeConvFast emerged as the top-performing model. It showcased outstanding performance across all key metrics in comparison to other state of the art models. As distance metric L1 distance is used for all models to set the threshold. Here the complete dataset is used and all models are equally optimized using optuna.

In a further evaluation on the simulated smart bridge dataset, ModeConvLaplace demonstrated significant superiority. It achieved the highest scores among all models. Here again L1 distance is used as distance metric to set the threshold and all modells are optimized using optuna equally.

\begin{table}[h]
\centering
\caption{Results for the complete simulated smart bridge dataset: ModeConv with L1 distance against other state of the art models}
\begin{tabular}{lcccc}
\toprule
Model & Precision (\%) & Recall (\%) & F1-Score (\%) & Accuracy (\%) \\
\midrule
MLP & 63.82 & 12.64 & 21.1 & 52.73 \\
VRAE & 73.87 & 21.54 & 33.34 & 56.98 \\
TCNAE & 88.09 & 45.02 & 59.48 & 69.50 \\
TGCN & 64.92 & 12.76 & 21.31 & 52.96 \\
STGCN & 78.39 & 23.95 & 36.57 & 58.67 \\
MTGNN & 67.57 & 15.19 & 24.78 & 53.95 \\
AGCRN & 65.87 & 13.18 & 21.95 & 53.19 \\
ModeConvFast & 97.4 & \textbf{90.7} & \textbf{93.9} & 66.5 \\
ModeConvLaplace & \textbf{99.4} & 86.5 & 92.5 & \textbf{86.8} \\
ChebConv & 99.3 & 55.4 & 71.1 & 73.5 \\
\bottomrule
\end{tabular}

\end{table}

Our findings highlight the efficacy of ModeConv variants in capturing complex spatiotemporal dependencies, especially evident in the ModeConvFast model on the Luxembourg dataset and the ModeConvLaplace model on the smart bridge dataset.

\section{Complexity reduction}
As demonstrated on both the Luxemburg and the Smart Bridge Dataset, the ModeConv Laplace model exhibits a runtime comparable to ChebConv due to a similar scheme of spectral graph convolution technique based on the normalized Laplacian, while ModeConv Fast has a significantely lower runtime due to its own convolutional filter (see ModeConv Method section).

To comprehend the ChebConv layer ~\cite{defferrard2017convolutional}, it is neccessary to grasp the concept of Spectral Graph Convolution (SGC) and subsequently, the influence of the Chebyshev polynomials $K$. Let $x(t): V \rightarrow \mathbb{R}^{N \times T}$ denote a signal within a multivariate time-series, defined at time step $t$ across the graph's nodes, where the hidden state is sequentially calculated for each timestamp $t$. The graph $G (n, e, w)$ comprises $n$ nodes, $e$ edges, and $w$ weights $\in \mathbb{R}^{N \times N}$, with the neighborhood $N_n = (e_1, \ldots, e_{d_n})^T$ for each node $n$ taken into consideration. 

In the case of SGC, $N_n$ is convolved with a degree-specific filter $f \in \mathbb{R}^{d_n}$ via $x(N_n)^T f$. Graphs can then be represented by the Graph Laplacian.

The Graph Laplacian $L$ is symmetric and thus diagonalizable. In $L = U\Lambda U^T$, $U \in \mathbb{R}^{N \times N}$ represents the orthogonal matrix of eigenvectors, and $\Lambda$ the diagonal matrix of corresponding eigenvalues of $L$.

This diagonalization facilitates defining the Graph Fourier Transform (GFT) $\hat{f}(t)$ of a signal $f(t) : V \rightarrow \mathbb{R}$ on the graph: $\hat{f} = U^T f$ and $f = U \hat{f}$. This GFT transforms a signal to the frequency or spectral domain. When a signal $x(t) : V \rightarrow \mathbb{R}^{N \times T}$ is convolved with a filter $F \in \mathbb{R}^{N \times N}$, this spectral filter can be defined as $x(t) * G F := U F U^T x(t)$.

Instead of employing the spectral convolutional filter as introduced, a polynomial approximation of order $K$ is utilized for the filter $F \approx \sum_{k=0}^{K} \theta_k \Lambda^k$ in case of the ChebConv layer, where $\theta \in \mathbb{R}^{K+1}$ is a vector of polynomial coefficients. The convolution then takes the following form:

\begin{align}
x(t) * G F &= U F U^T x(t) \approx U \sum_{k=0}^{K} \theta_k \Lambda^k U^T x(t) \
&= \sum_{k=0}^{K} \theta_k U \Lambda^k U^T x(t)
\end{align}

Employing Chebyshev polynomials offers two distinct advantages: $T_k(x)$ can be efficiently computed due to its recursive definition, and $U$ and $U^T$ are no longer necessary in the convolution:
\begin{align}
x(t) * G F &= U F U^T x(t) \\
&\approx U \sum_{k=0}^{K} \theta_k T_k(\Lambda) U^T x(t) \\
&= \sum_{k=0}^{K} \theta_k U T_k(\Lambda) U^T x(t) \\
&= \sum_{k=0}^{K} \theta_k T_k(U^T \Lambda U) x(t) \\
&= \sum_{k=0}^{K} \theta_k T_k(L) x(t)
\end{align}

Regarding the utilization of Chebyshev polynomials, the time complexity is denoted by $O(K|E|)$, where $K$ denotes the order of the Chebyshev polynomial, and $|E|$ represents the number of edges in the graph~\cite{defferrard2017convolutional}. Considering $|E|$ as the number of edges in the graph, the time complexity for ChebConv becomes $O(K|E|)$. For a fully connected graph with $n$ nodes, we have:

\begin{equation}
|E| = \binom{n}{2} = \frac{n \times (n - 1)}{2}
\end{equation}

The order $K$ of the Chebyshev polynomial thus plays a significant role in the performance of the ChebConv layer. Increasing the order of the polynomial generally leads to higher approximation accuracy of the spectral filters. A higher-order polynomial can capture more intricate spectral characteristics, allowing for finer-grained analysis of the graph signals. However, this comes at the cost of increased computational complexity. As $K$ grows, the computational burden of computing and applying the polynomial approximation also increases, potentially leading to longer training and inference times, especially in large-scale graph datasets. While a higher-order polynomial offers improved accuracy in capturing spectral features, it may also introduce computational overhead. Therefore, practitioners often need to balance between the desired level of accuracy and the computational resources available.

On the other hand, the time complexity of ModeConv Fast involves computing the Singular Value Decomposition (SVD) of an $m \times n$ matrix, where $m$ is the number of eigenmodes and $n$ is the number of sensors. 
Assuming that $m$ represents the number of chosen eigenmodes and $n$ represents the number of sensors, based on the LAPACK implementation used by numpy the time complexity for computing the SVD of the signal $x$ is then given by~\cite{henry19928}:
\begin{equation}
    T_{\text{SVD}} = 
    \begin{cases} 
    O(mn^2) & \text{if } m \geq n \\
    O(n^3) & \text{if } m < n 
    \end{cases}
\end{equation}

Therefore, for a fully connected graph with $n$ nodes, ChebConv has a higher time complexity than ModeConvFast when $K = 5$, which was optimized during training. This comparison is highly dependent on the specified parameters and the structure of the graph. Notably, as the graph densifies with more edge connections, the efficiency of ModeConv Fast accentuates significantly. This is attributed to its adeptness at reducing the matrix size in accordance with the number of eigenmodes. Consequently, in scenarios in which the graph has a large number of edges with a significantly smaller number of eigenmodes compared to sensors, the efficiency gains of ModeConv Fast become increasingly visible.
In the conducted additional profiling study, the results indicate that the ModeConv Fast exhibits different speed profiles compared to ChebConv. While the forward pass ({\tt call\_impl}) of the model in ModeConv Fast becomes faster, the preprocessing in the {\tt process\_batch} takes longer. Specifically, the computation of covariances and the execution of SVD contribute to this increased time requirement. This suggests that ModeConv Fast requires additional time for preprocessing to generate the necessary weight matrices and similarity values, whereas ChebConv creates the Laplacian matrix during the forward pass, leading to a lower percentage of total time for preprocessing. Precomputation of  preprocessing steps allowed to further enhance the efficiency of ModeConv Fast, which has been implemented. %

\section{Conclusion}
ModeConv offers a novel approach that integrates graph-based representation, covariance and PSD computation, modal decomposition, and graph convolutions for continuous condition monitoring of structures. ModeConv is build upon existing techniques and concepts such as Spectral Graph Convolutions and Fourier/Laplace transforms as base. As novelty, it uses a novel complex convolutional filter to learn the spatial and temporal patterns in the extracted features and leverages the modal coordinates as inputs to this convolution. This combination enables the detection of subtle structural changes and facilitates accurate damage detection. %
It further has potential applications in a wide range of domains beyond structural health monitoring, such as speech recognition, audio data, mechanics, quantum physics or simplified in anything that vibrates.

This paper introduces a novel extension to Spectral Graph Convolutions by modulating natural modes and retaining only the relevant natural frequencies while eliminating irrelevant frequencies. The main advantages claimed for ModeConv are better results on anomaly detection tasks related to physical material properties and improved runtime and parameter efficiency.
 
In reference to the first claim, the ModeConv model, particularly the ModeConvFast variant, mostly achieved higher F1 scores, precision, and recall compared to other models (ChebConv, MtGNN, AGCRN) on the both large-scale datasets. This indicates that ModeConv performs well in detecting anomalies in sensor data based on physical material properties. In reference to the second claim, the ModeConvFast model achieved the best performance in terms of F1 score, balanced accuracy, and AUC on the Luxemburg dataset, while also having the lowest training time per epoch. 

Additionally, a comparative study was conducted to examine the impact of different components of the ModeConv layer. The results showed that the combination of using two linear layers instead of Power Spectral Density calculation and utilizing the Laplacian instead of SVD in the Convolutional block resulted in similar performances of all ModeConv variants. 

\section{Acknowledgements}
We thank the German Federal Highway Research Institute (Bundesanstalt für Straßenwesen = BASt) for providing us with the smart bridge dataset as well as the simulated damages dataset in cooperation with Prof. Freundt and IB Freundt. Furthermore we thank the German Federal Ministry for Digital and Transport (BMDV) for financing the P-BIM project. Last but not least we are expressing our heartfelt thanks to Prof. Maas and Khatereh Dakhili from the University of Luxembourg for sharing the \emph{Luxemburg dataset} with us.

\bibliographystyle{ACM-Reference-Format}
\bibliography{main}

\end{document}